\documentclass[11pt]{article}

\usepackage{ACL2023}

\usepackage{times}
\usepackage{latexsym}
\usepackage{subcaption}
\usepackage{graphics}
\usepackage{graphicx}
\usepackage{hyperref}
\usepackage{dsfont}
\usepackage{amsmath}
\usepackage{xspace}
\usepackage{enumitem}
\usepackage{amsmath}
\usepackage{amsfonts}
\usepackage[normalem]{ulem}
\usepackage{booktabs}
\usepackage{algorithm}% http://ctan.org/pkg/algorithm
\usepackage{algpseudocode}% http://ctan.org/pkg/algorithmicx
\usepackage[T1]{fontenc}
\usepackage[utf8]{inputenc}

\usepackage{microtype}
\usepackage{comment}
\usepackage{todonotes}
\usepackage{caption}
\usepackage{subcaption}

\newcommand{\secref}[1]{\S\ref{#1}}
\newcommand{\clues}{{\fontfamily{lmtt}\selectfont CLUES}\xspace}

\newcommand{\cluesreal}{{\fontfamily{lmtt}\selectfont \clues-Real}\xspace}

\newcommand{\exent}{{\fontfamily{lmtt}\selectfont ExEnt}\xspace}

\newcommand{\nalla}{{\fontfamily{lmtt}\selectfont TALC}\xspace}
\newcommand{\Thead}[1]{\textbf{\textsc{#1}}}
\definecolor{LightCyan}{RGB}{172,204,186}

\title{
Leveraging Multiple Teachers for Test-Time Adaptation of Language-Guided Classifiers
}
\author{
Kangda Wei$^{1}$ \qquad  Sayan Ghosh$^{2}$ \quad Rakesh R. Menon$^{2}$ \qquad  Shashank Srivastava$^{2}$\\
$^{1}$Department of Computer Science and Engineering, Texas A\&M University \\
$^{2}$Department of Computer Science, UNC Chapel Hill \\
\texttt{kangda@tamu.edu} \quad 
\texttt{\{sayghosh, rrmenon, ssrivastava\}@cs.unc.edu}
}

\begin{document}
\maketitle

\begin{abstract}
    Recent approaches have explored language-guided classifiers capable of classifying examples from novel tasks when provided with task-specific natural language explanations, instructions or prompts \cite{sanh2022multitask,menon-etal-2022-clues}. 
    While these classifiers can generalize in zero-shot settings, their task performance often varies substantially between different language explanations in unpredictable ways \cite{lu2022fantastically, gonen2022demystifying}. Also, current approaches fail to leverage unlabeled examples that may be available in many scenarios. Here, we introduce \nalla, a framework that uses data programming to adapt a language-guided classifier for a new task during inference when provided with explanations from multiple teachers and unlabeled test examples. Our results show that \nalla consistently outperforms a competitive baseline from prior work by an impressive 9.3\% (relative improvement). Further, we demonstrate the robustness of \nalla to variations in the quality and quantity of provided explanations, highlighting its potential in scenarios where learning from multiple teachers or a crowd is involved. Our code is available at: \url{https://github.com/WeiKangda/TALC.git}. 
\end{abstract}

\section{Introduction}

\begin{figure}[!t]
    \centering
    \includegraphics[scale=0.63]{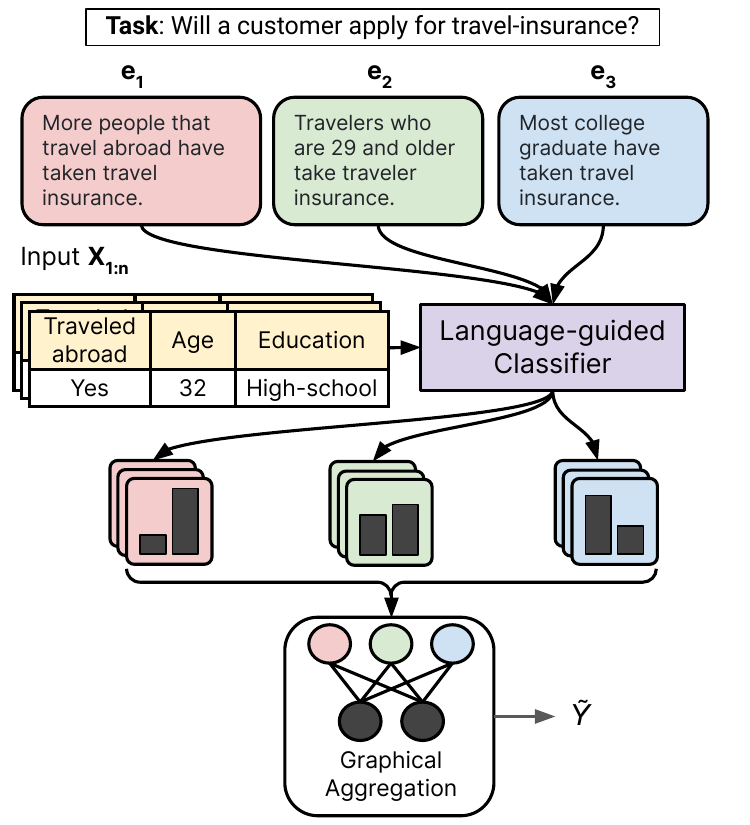}
    \caption{
    \textbf{\nalla} leverages data programming to perform test-time adaptation of a language-guided classifier.
    Natural language explanations ($E= \{e_1, e_2, e_3 \}$) provided by multiple teachers and unlabeled examples ($X_{1:n}$) for a new task are fed to a language-guided classifier pair-wise resulting in multiple pseudo-labels for the unlabeled examples. 
    \nalla uses a graphical aggregation to weigh the pseudo-labels from different explanations to decide the final predicted label ($\hat{Y}$). \nalla is highly flexible in aggregating labels as it can conceptually consider a broad variety of factors, such as the complexity of explanations, consistency between explanation predictions, identity of the explanation provider, etc.} 
    \vspace*{-0.5cm}
    \label{fig:intro}
\end{figure}

Inductive learning from examples has underpinned many successful machine learning applications. However, classifiers trained solely from labeled examples often struggle to generalize in scenarios with limited labeled data. In contrast, humans can learn new concepts through natural language conversations~\cite{chopra2019first, 10.2307/j.ctvjsf4jc}. Inspired by this phenomenon, recent approaches use natural language explanations, instructions, and prompts to train \emph{language-guided classifiers} \cite{srivastava-etal-2017-joint, andreas-etal-2018-learning, murty-etal-2020-expbert,  Wang*2020Learning, DBLP:journals/corr/abs-2005-00806}.
While these classifiers can perform zero-shot classification, they have several limitations. Firstly, they lack a principled strategy for weighing language supervision from multiple sources (or teachers). 

Secondly, they fail to utilize the available unlabeled data for a new task during inference.  Additionally, the impact of the quality of sources, and the inclusion of low-quality explanations, remains largely unexplored.

To address these limitations, we present \nalla (\textbf{T}est-time \textbf{A}daptation of \textbf{L}anguage-guided \textbf{C}lassifiers), a framework for adapting language-guided classifier on novel tasks during inference, also known as test-time adaptation. \nalla assumes a priori access to the entire test set (unlabeled samples) of the novel task and the task-specific explanations, which aligns with real-world situations, such as developing a product-category classifier for an e-commerce platform. 
In the context of \nalla, the multiple explanations available for each task are considered as distinct supervisory signals. Leveraging the power of data programming~\cite{Ratner_2018}, \nalla aggregates and combines the supervision provided by these multiple explanations effectively.

Figure~\ref{fig:intro} illustrates the \nalla framework. \nalla uses a subset of the test data, called the adaptation set, for adapting the language-guided classifier. For each pair of explanation and test example in the adaptation set, a pseudo-label is generated using the base language-guided classifier. \nalla learns a label aggregator on the pseudo-labels generated for the adaptation set using EM (\secref{sec:talc}). The label aggregator is trained to consider the contribution of each explanation during adaptation, thus, in principle, allowing it to weigh different sources of language supervision. Finally, \nalla uses the learned aggregator over the entire test set to obtain final predictions for test set examples. 

We evaluate \nalla on six classification tasks from the \cluesreal dataset~\cite{menon-etal-2022-clues}, where each task is paired with natural language explanations (\secref{sec:exp_and_results}). \nalla outperforms strong baselines by 3.3\% on average (absolute). Through qualitative and quantitative analysis, we investigate \nalla's robustness with respect to the size of the adaptation set, number of explanations, and explanation quality. 
In the subsequent sections, we describe \nalla in detail (\secref{sec:talc}), present experimental results and analysis (\secref{sec:exp_and_results}), and conclude by discussing our contributions, limitations, and ethical considerations. Our contributions are:
\begin{itemize}[noitemsep, topsep=0pt, leftmargin=*]
    \item We introduce \nalla, a test-time adaptation framework, that uses label aggregation to improve language-guided classifiers.
    \item We demonstrate the effectiveness of \nalla on multiple real-world classification tasks from \cluesreal~\cite{menon-etal-2022-clues}. 
    \item We present comprehensive analyses to evaluate the robustness of \nalla w.r.t. the quantity and quality of explanations.
\end{itemize}

\section{Related Work}
\paragraph{Learning From Language}
Using natural language explanations to inform or train classifiers has garnered significant interest in recent years \cite{7524b90aadb74d5abcd045aaf8407779, srivastava-etal-2017-joint, hancock-etal-2018-training, murty-etal-2020-expbert}. While \citet{murty-etal-2020-expbert} enhance supervised BERT models \cite{devlin-etal-2019-bert} for relation extraction tasks, other approaches employ language explanations for few-shot learning. For instance, \citet{hancock-etal-2018-training} convert explanations to labeling functions via semantic parsing, leveraging unlabeled data for weak labels. More recently, \citet{menon-etal-2022-clues} utilize natural language explanations in an entailment-based model for classification decisions.

\paragraph{Test-time Adaptation}
Test-time adaptation has been extensively studied in computer vision by employing batch-normalization statistics \cite{https://doi.org/10.48550/arxiv.2006.10963, https://doi.org/10.48550/arxiv.2112.02355, https://doi.org/10.48550/arxiv.2006.16971}, test-time entropy minimization \cite{https://doi.org/10.48550/arxiv.2006.10726, https://doi.org/10.48550/arxiv.2110.10232}, prediction consistency maximization over augmentations \cite{https://doi.org/10.48550/arxiv.2110.09506}, and classifier adjustment \cite{NEURIPS2021_1415fe9f}.
In the realm of natural language processing, \citet{banerjee-etal-2021-self} explore test-time adaptation for question-answering using self-supervision. In contrast, we introduce a new test-time adaptation approach that leverages data programming to adapt a base language-guided classifier during inference by utilizing natural language explanations. 

\paragraph{Data Programming}
Data programming \cite{Ratner_2017} employs a combination of multiple labeling functions and generative models to create probabilistic training labels for unlabeled datasets.
Prior work \cite{Ratner_2018, hancock-etal-2018-training} has demonstrated successful applications of this paradigm to create systems that allow users to label large datasets programmatically. 
Here, we repurpose data programming in the test-time adaptation setting to improve classifiers on unseen tasks.

\section{\nalla}
\label{sec:talc}
In this section, we present the details of our framework, \nalla. 
\nalla leverages data programming to adapt a base natural language explanation-guided classifier on a novel task during inference. 

\paragraph{Problem Setup.} We assume a language-guided classifier, $\mathcal{M}_{LC}$, which can take an explanation $e$ from a teacher and example $X$ to predict a label $\mathcal{M}_{LC}(X, e)$ . A language-guided classifier refers to a classifier that utilizes one or more natural language explanations to make predictions. Our objective is to make predictions for a batch of test samples, represented as $\{X_{test}, Y_{test}\}_{1:n}$, where $Y_{test}$ represents the unobserved ground-truth labels corresponding to the samples in $X_{test}$, and $n$ denotes the number of samples. During test-time adaptation, our aim is to effectively adapt the classifier to the specific task at hand and infer the true labels for $X_{test}$. Existing methods for test-time adaptation typically assume an online setting, where examples are processed one at a time \cite{sun19ttt, banerjee-etal-2021-self}. In contrast, we assume a priori access to the entire test set of the task. This assumption allows us to leverage the empirical distribution of the unlabeled data for semi-supervised learning. 

Our setting aligns with real-world scenarios, such as developing a product-category classifier for an e-commerce platform, where the complete database of products (including the test set) is known in advance.
For situations where test samples are observed one at a time, it is still possible to utilize \nalla for adapting a base classifier. This involves a "warm-up" phase, where the base classifier is used off-the-shelf for a few samples, followed by adaptation using \nalla. While this usage scenario is not the primary focus of our work, we provide a description of how \nalla can be employed in such cases in Appendix~\secref{sec:talc_usage} for brevity.

\paragraph{Overview.} As depicted in Figure~\ref{fig:intro}, for a new task $T_{new}$, we are provided with $m$ natural language explanations $E = \{e_1, e_2, \ldots, e_m\}$, and a set of examples $\{ X_i \in X_{test}\}$. To generate the classifier outputs, we iterate through each explanation $e_j$ for every example $X_i$, and compute $M_{ij} := \mathcal{M}_{LC}(X_i, e_j)$. 
This yields a labeling matrix $M$ with a shape of $n \times m$. Next, we introduce a test-time adaptation procedure: \nalla, to compute the final labels $\tilde{Y}$ utilizing the $M$. This procedure essentially implements a function $f : M \in \mathds{R}^{n \times m} \rightarrow \tilde{Y}\in \mathds{R}^n$, which we describe in the rest of this section. 

\begin{algorithm}
  \caption{\nalla}\label{alg: test-time adaptation}
  \hspace*{\algorithmicindent} \textbf{Inputs:} Language-guided classifier $\mathcal{M}_{LC}$, test set $X_{test}$, task explanations $E$, adaptation ratio $\alpha$ 
  \begin{algorithmic}[1]
  \State $N_{adapt} \leftarrow \alpha \times |X_{test}|$
    \State $X_{test}^{adapt} \leftarrow X_{test}[:N_{adapt}]$
    \State $X_{test}^{held-out} \leftarrow X_{test}[N_{adapt}:]$
    \State Train the label aggregator $\mathcal{L}^{\textrm{agg}}_w$ on $X_{test}^{adapt}$ 
    $\hat{w} \leftarrow \underset{w}{\textrm{argmax}}\; P_w(X, E; \mathcal{M}_{LC}$ (using EM)
    \State Infer $\tilde{Y}_{\nalla}$ for $X_{test}$ using the learned $\hat{w}$.  $\tilde{Y}_{\nalla} := \underset{Y}{\textrm{argmax}}\; P_{\hat{w}}(Y| X_{test}, E, \mathcal{M}_{LC})$ 
  \end{algorithmic}
  \hspace*{\algorithmicindent} \textbf{return} $\tilde{Y}_{\nalla}$
\end{algorithm}

\paragraph{Test-time Adaptation.}
The objective of \nalla is to adapt the language-guided classifier, $\mathcal{M}_{LC}$, on a novel task, $T_{new}$ during inference. We illustrate the adaptation procedure in Algorithm~\ref{alg: test-time adaptation}.
First, we split the test set into two disjoint sets - the adaptation set and the held-out test. The adaptation set is utilized by \nalla to adapt $\mathcal{M}_{LC}$. The proportion of the test set that forms the adaptation set is defined by an adaptation ratio, $\alpha \in [0,1]$, defined as
$\alpha = \frac{|\text{adaptation set}|}{|\text{test set}|}$
We also partition the labeling matrix $M$ into $M^{adapt}$ and $M^{held-out}$ by choosing the rows corresponding to samples in the adaptation set and held-out set, respectively.

To model the dependence between the (latent) inferred labels and $M^{adapt}$, we use data programming techniques \cite{ratner2019training} to train a label aggregator, $\mathcal{L}^{\textrm{agg}}_w$, with task-specific parameters $w$. We use the learned parameters (which correspond to weights learned for each explanation) to aggregate predictions in $M$ (both $M^{adapt}$ and $M^{held-out}$) and infer the labels, $\tilde{Y}_{\nalla}$. 

\paragraph{Label Aggregator.}
The label aggregator is a graphical model that defines the joint probability of explanations $E$, examples $X$ and latent (true) labels $Y$ for a given language-guided classifier as:

\vspace{-0.5cm}
\begin{equation}
 \label{eqn:graphical-model}
  P(X, E, Y ; \mathcal{M}_{LC}) \propto \exp{ w^T \phi(X, E, Y, \mathcal{M}_{LC})}   
\end{equation}

Here, $\phi$ is a feature-vector of features that can be computed using $X$, $E$, $Y$ and $\mathcal{M}_{LC}$ and $w$ is a weight vector corresponding to each of those features. In general, this can subsume a very broad range of features, including features that can indicate the complexity of an explanation, or its provenance~\footnote{So, for example, the aggregator can automatically learn to lean more/less on complex explanations, or trust explanations from specific sources more than from others}. We also note that in particular, since the labeling matrix $M$ is computed from $X$, $E$ and $\mathcal{M}_{LC}$, $\phi$ can include features that depend on $M$ and $Y$. 
For simplicity, our instantiation incorporates the labeling rates of each explanation (how frequently an explanation assigns a label\footnote{The pseudo-label corresponding to an explanation can either be a class label,  or a special label, $y_{abstain}$ (e.g. if an explanation does not apply for an example)}) and the correlations between the pseudo-labels from different explanations to estimate the accuracies of each individual explanation in an unsupervised manner. 
Specifically, the label aggregator is defined in terms of two types of features: accuracy ($\phi^{Acc} \in \mathbb{R}^{n\times m}$) and propensity ($\phi^{Prop}\in \mathbb{R}^{n\times m}$). Each value in $\phi^{Acc}$ and $\phi^{Prop}$ is defined as:
    \begin{equation}
    \label{eq: accuracy}
        \phi_{i,j}^{Acc}(M, Y) = \mathds{1}\{M_{i,j} = y_i\}
    \end{equation}
    \vspace{-0.65cm}
    \begin{equation}
    \label{eq: propensity}
        \phi_{i,j}^{Prop}(M, Y) = \mathds{1}\{M_{i,j} \ne y_{abstain}\}
    \end{equation}
    where $y_i$ is the label for $i^{th}$ sample and  $y_{abstain}$ is a special label that denotes $\mathcal{M}_{LC}$ has abstained from predicting a label based on the $j^{th}$ explanation.
The accuracy factor fires if the inferred label ($Y_{i}$) for an unlabeled example $X_i$ matches the predicted label from an explanation $j$. The propensity factor fires whenever the classifier doesn't abstain from predicting a label from an explanation.

Since here we only define two types of features for each explanation, $w \in \mathds{R}^{2m}$ is a learnable vector corresponding to weights for the accuracy factor and propensity factor for each explanation. The weights are learned by maximizing the log-likelihood $\log P(X, E) = \log \sum_Y P_w(X, E, Y)$ using the expectation-maximization (EM) algorithm (since we don't have ground-truth labels for $Y$ at test-time).
We compute the MAP estimate $\tilde{Y}_{\nalla} := \underset{Y}{\textrm{argmax}} P_{\hat{w}}(Y | X, E)$ using Gibbs sampling to predict the final labels. Note that while we learn the weights $\hat{w}$ on the adaptation set (line 4 in Algorithm 1), the learned weights are used to aggregate predictions in both the adaptation and the held-out examples, 
to predict the labels, $\tilde{Y}_{\nalla}$ (line 5 in Algorithm 1). We implement the label aggregator using Snorkel-Metal\footnote{\url{https://www.snorkel.org/}}~\cite{ratner2018snorkel}. Appendix~\secref{sec:app_hyp} provides  task-specific details of the label aggregator training. 

\section{Experiment and Analysis}
\label{sec:exp_and_results}
In this section, we evaluate the zero-shot adaptation performance of \nalla on classification tasks, followed by a detailed analysis of \nalla's robustness.

\subsection{Data}
\label{sec: data}
We assess the performance of \nalla on real-world classification tasks from the \clues~\cite{menon-etal-2022-clues} benchmark. Out of the sixteen real-world tasks in the test split of \clues, we focus on six tasks for evaluation due to the limited number of test samples ($<10$) in the remaining tasks, which restricts their suitability for test-time adaptation. Figure~\ref{fig:intro} presents an illustrative example showcasing the nature of these tasks and provides examples of the corresponding natural language explanations. Appendix~\secref{sec:app_data} provides further details regarding the six tasks selected for evaluation.

We utilize the \exent model \cite{menon-etal-2022-clues} as the base language-guided classifier ($\mathcal{M}_{LC}$) in alignment with our choice of the \clues dataset\footnote{At the time of writing, this is the best model on \clues with publicly available code.}. The \exent model leverages textual entailment to establish the correspondence between explanations and tabular inputs, enabling label predictions. To aggregate predictions from multiple explanations, \exent adopts a mean-pooling strategy, aggregating the predictions obtained from each explanation-input pair to derive the final label. It's important to note that \exent is not trained for abstention, meaning it always assigns a label regardless of the quality of the explanations. In \secref{sec:analysis}, we further explore the scenario of abstention, which we consider a more realistic use case for language-guided classifiers.

\subsection{Baseline and Evaluation Metrics}
\label{sec: baseline}

We compare \nalla against the following baselines:
\begin{enumerate}[noitemsep, topsep=0pt, leftmargin=*]
    \item \underline{\exent}: This refers to the base \exent model~\cite{menon-etal-2022-clues} that has been trained on real-world training tasks from the \clues dataset.

    \item \underline{\exent-MV}: For each example $X_i$, we generate a set of pseudo-labels corresponding to each of the $m$ task explanations. The final predicted label is determined by selecting the label that appears most frequently among the $m$ pseudo-labels (\textit{majority vote}). Unlike \exent, which uses mean-pooling for aggregation, \exent-MV applies a mode-pooling operation.

    \item \underline{\exent-FT}: Similar to our approach of fine-tuning \nalla with the predicted labels from the label aggregator, $\mathcal{L}^{\textrm{agg}}_w$, we also include a self-training baseline approach by \textit{fine-tuning} \exent. This involves utilizing \exent's own predictions as labels on the adaptation set.
\end{enumerate}
We use classification accuracy as the evaluation metric to compare the utility of different methods.

\subsection{Results}\label{sec:results}
\begin{table*}[t!]
\small
\begin{center}
\begin{tabular}{ lrr|rr|rr} 
 \toprule
 & \multicolumn{2}{c|}{\begin{tabular}{@{}c@{}}\textbf{Non-Adaptation}\\\textbf{Baselines}\end{tabular}} 
 & \multicolumn{2}{c|}{\textbf{Adaptation Ratio 0.5}} & \multicolumn{2}{c}{\textbf{Adaptation Ratio 1.0}} \\

\textbf{\textbf{Tasks}} & \exent & \begin{tabular}{@{}c@{}}\exent\\MV\end{tabular} 
& \begin{tabular}{@{}c@{}}\exent-FT\end{tabular} & \nalla & \begin{tabular}{@{}c@{}}\exent-FT\end{tabular} & \nalla\\

 \midrule
 banknote-authentication & $46.9$ & $48.4$  & $45.1_{(0.5)}$ & $49.5_{(0.0)}$ & $45.2_{(0.2)}$ & $\mathbf{49.7_{(0.3)}}$ \\ % & $56.7$
 tic-tac-toe-endgame & $32.8$ & $32.3$  & $32.3_{(0.0)}$ & $\mathbf{41.1_{(0.0)}}$ & $32.3_{(0.0)}$ & $\mathbf{41.1_{(0.0)}}$ \\ % & $65.1$
 car-evaluation & $10.7$  & $\mathbf{17.6}$  & $4.6_{(2.3)}$ & $14.1_{(3.4)}$ & $3.8_{(0.4)}$ & $16.5_{(0.0)}$\\ % & $42.8$
contraceptive-choice & $42.7$ & $43.7$  & $43.4_{(0.0)}$ & $\mathbf{44.0_{(0.7)}}$ & $43.4_{(0.0)}$ & $43.8_{(0.3)}$  \\ % & $46.8$
 indian-liver-patient & $48.7$  & $40.0$   & $54.5_{(7.8)}$ & $44.3_{(2.8)}$ &$\mathbf{62.0_{(4.6)}}$ & $47.8_{(0.0)}$ \\ % & $54.8$
 travel-insurance & $31.9$ & $33.9$ & $\mathbf{34.2_{(0.0)}}$ & $\mathbf{34.2_{(0.0)}}$ & $\mathbf{34.2_{(0.0)}}$ & $\mathbf{34.2_{(0.0)}}$ \\ %  & $77.1$
 \midrule
 \textbf{Average} & $35.6$ & $36.0$  & $35.7$ & $37.9$ & $36.8$ & $\mathbf{38.9}$ \\ % & $57.2$
\bottomrule
\end{tabular}
\end{center}
\vspace{-0.1in}
\caption{Comparison of zero-shot accuracies (higher is better) between non-adaptation-based \exent baselines, \exent-FT, and our proposed method, \nalla, on the 6 different tasks from \cluesreal. We report the mean and standard deviation for the accuracy across three runs for adaptation-based methods.
 The numbers in \textbf{bold} indicate the best accuracies across methods.
}
\vspace{-0.1in}
\label{tab:main-results}
\end{table*}
Table~\ref{tab:main-results} shows the zero-shot classification accuracy of \nalla and the baselines on the six evaluation tasks. The findings reveal several key insights. Firstly, we observe that majority voting (\exent-MV) performs better than vanilla \exent on average across the six tasks. Secondly, fine-tuning \exent on its own predictions (\exent-FT) results in better zero-shot accuracies than the base \exent model, demonstrating the value of self-training on unlabeled data. Furthermore, the performance of \exent-FT increases with an increase in the amount of test data used for adaptation ($35.7 \rightarrow 36.8$ as we increase the adaptation ratio from $0.5 \rightarrow 1.0$). 

We note that \nalla obtains better performance on average across all evaluation tasks compared to the three baselines. Specifically, \nalla improves the accuracy by around $3.3\%$ on average (absolute) over the state-of-the-art \exent model. In fact, both \nalla variants, at adaptation ratio $0.5$ and $1.0$, perform better than \exent on all tasks except for indian-liver-patient. The utilization of the label aggregator in \nalla results the biggest improvement ($\sim25\%$ relative) on the tic-tac-toe-endgame task. We attribute this improvement to the label aggregator's ability to give higher weightage to high-quality explanations, resulting in more accurate predictions.
For the tasks where the performance of \nalla is close for $\alpha=0.5$ and $\alpha=1.0$ in Table 1, we observed that the aggregation weights for each explanation learned by the data programming framework are roughly similar for the two settings. As a result, the aggregation over the pseudo labels in the labeling matrix produces similar final predictions and hence similar accuracies.

\begin{table*}[t!]
\small
\begin{center}
\begin{tabular}{ lrr|rr} 
 \toprule
 & \multicolumn{2}{c|}{
 \textbf{Non-Adaptation Baselines}
 } & \multicolumn{2}{c}{\textbf{Adaptation Methods}}\\
\textbf{\textbf{Tasks}} & \exent-A & \begin{tabular}{@{}c@{}}\exent\\MV-A\end{tabular} 
& \begin{tabular}{@{}c@{}}\exent-FT-A\end{tabular} & \nalla-A \\
 \midrule
 banknote-authentication & $8.0$ & $36.7$  & $27.1_{(2.5)}$ & $\mathbf{53.9_{(0.1)}}$ \\
 tic-tac-toe-endgame & $2.6$ & $30.2$  & $32.3_{(0.1)}$ & $\mathbf{32.4_{(0.4)}}$\\ 
 car-evaluation & $2.3$  & $2.6$  & $\mathbf{14.8_{(1.8)}}$ & $13.3_{(0.8)}$ \\ 
contraceptive-choice & $22.7$ & $\mathbf{32.8}$  & $19.3_{(2.7)}$ & $31.6_{(1.3)}$ \\ 
 indian-liver-patient & $36.5$  & $34.7$   & $30.1_{(0.2)}$ & $\mathbf{40.2_{(0.9)}}$\\ 
 travel-insurance & $15.3$ & $\mathbf{24.6}$ & $15.2_{(2.6)}$ & $23.4_{(0.0)}$ \\
 \midrule
 \textbf{Average} & $14.6$ & $26.9$  & $23.1$ & $\mathbf{32.5}$\\
\bottomrule
\end{tabular}
\end{center}
\caption{Comparison of zero-shot accuracies between \nalla and the baselines when allowing the \exent model to abstain from making a
prediction (the modified model is denoted as \exent-A). `A' stands for `Abstention' for all the models in the table.
For the adaptation methods (\exent-FT, \nalla), we report mean and standard deviation across 9 adaptation
ratios (0.2 to 1.0). Numbers in \textbf{bold} denote the best accuracies across methods.
}
\vspace{-0.1in}
\label{tab:abstention-results}
\end{table*}
\subsection{Analysis}
\label{sec:analysis}
\paragraph{Abstention.}
Our previous experimental results treat labels from individual explanations the same, irrespective of the confidence of the model in its predictions on those examples. This is because the base-language classifier used in experiments, \exent, always chooses a label during inference rather than performing selective predictions. However, the \nalla framework allows for differential modeling of abstentions, where a model can choose to refrain from assigning a class label if the explanation does not apply to the example.  To explore this, we design a variant of \exent, referred to as \exent-A, that can refrain from assigning a class label during inference. This is straightforward since \exent is based on NLI \cite{menon-etal-2022-clues}, where a neutral label can be naturally mapped to an abstention. We train \exent-A on the same tasks as \exent with the modification of having `abstain' as an additional class label for each task. 

Table~\ref{tab:abstention-results} shows the results of \nalla-A, and the baselines when abstention is allowed (`-A' denotes abstention). We find that \nalla-A achieves the best overall accuracy. More importantly, comparing Table~\ref{tab:main-results} and Table~\ref{tab:abstention-results}, we observe that \nalla has a smaller drop in performance in comparison to \exent and \exent-FT suggesting that \nalla is better at adapting to multiple teachers even when certain teachers choose to abstain from prediction.

\paragraph{Effect of adaptation set size.}
\begin{figure}[!h]
    \centering
    \par
    \vspace{-0.5cm}
    \includegraphics[scale=0.53]{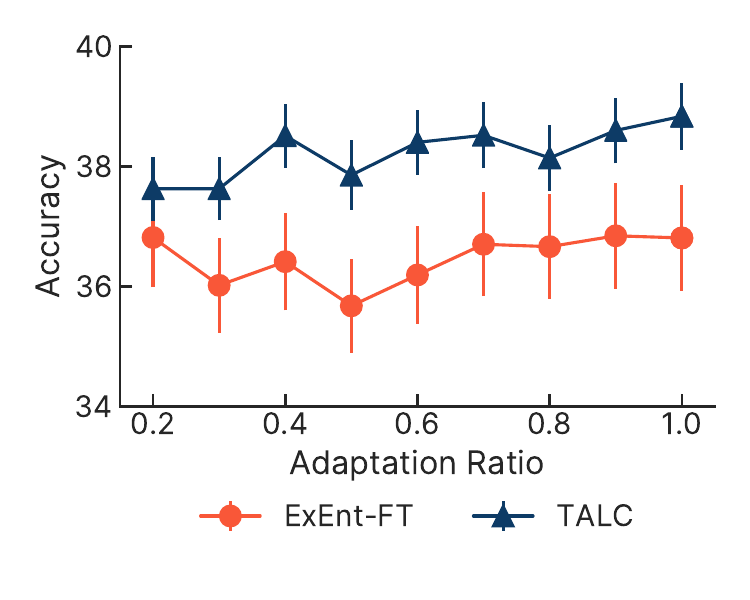} % result_various_explanations.pdf
    \vspace{-0.6cm}
    \caption{Accuracy (averaged over 6 tasks) of \exent-FT and \nalla when training label aggregator with different adaption set sizes. Overall, increasing the adaptation ratio does not impact performance of \exent-FT, but improves performance of \nalla.
 }
    \label{fig:different adaptations}
    \vspace{-0.3cm}
\end{figure}

We analyze the performance of \exent-FT and \nalla by varying size of the adaptation set. Specifically, we vary the adaptation ratio, $\alpha$, from $0.2$ to $1.0$ (in increments of 0.1) for all six evaluation tasks.  
\footnote{The results of \nalla and \exent-FT on each individual task with different adaptation ratios can be found in Appendix~\secref{sec:app_noabst_results}.}

Intuitively, we expect that the accuracy of \exent-FT and \nalla to improve with increase in adaptation ratio. However, we empirically observe that the performance of \exent-FT fluctuates with change in $\alpha$ and does not show a consistent trend of improvement as shown in Figure~\ref{fig:different adaptations}. Meanwhile, as shown in Figure~\ref{fig:different adaptations}, we observe that a larger adaptation set enhances the performance of \nalla from $37.6\% \rightarrow 38.8\%$ as $\alpha$ increases from $0.2 \rightarrow 1.0$. 

\begin{figure}[!h]
    \centering
    \includegraphics[scale=0.53]{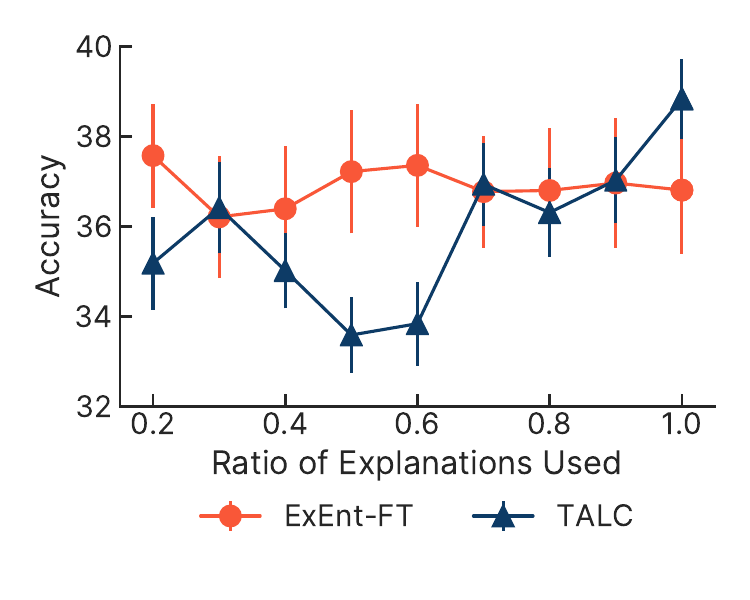}
    \vspace{-0.4cm}
    \caption{Results for \exent-FT and \nalla when varying the number of explanations used for training the label aggregator. Results are averaged over the six evaluation tasks. With increase in number of explanations, the accuracies using \nalla improve while performance of \exent-FT is not affected.
 }
    \label{fig:different explanations}
\end{figure}
\paragraph{Robustness to number of explanations.}
    Next, we analyze the robustness of \exent-FT and \nalla to changes in the number of explanations provided for adaptation on the new task.
We will refer to the fraction of explanations used for adaptation as the $\textrm{explanation ratio} = \frac{\text{\#~of~available~expls}}{\text{\#~of~all~expls}}$. Specifically, we vary the explanation ratio from $0.2$ to $1.0$, by randomly choosing explanations without replacement, when training the \exent-FT model and the label aggregator in \nalla. We keep the adaptation ratio ($\alpha$) fixed at 1.0 for this analysis.

Figure~\ref{fig:different explanations} shows the variation in performance of \exent-FT and \nalla with changes in the explanation ratio averaged over the six evaluation tasks.
The accuracy of \nalla drops when increasing the explanation ratio as $0.3 \rightarrow 0.5$, buts shows a consistent increasing trend (from $33.5\% \rightarrow 38.8\%$) when increasing the explanation ratio from $0.5 \rightarrow 1.0$. In contrast, the performance of \exent-FT fluctuates as the number of available explanations changes. This shows that \nalla is comparatively more sensitive to the number of explanations used for adaptation.

\paragraph{Robustness to quality of explanations.}
Here we analyze the role of explanation quality on the performance of \nalla. 
However, quantifying the quality of explanations in the absence of annotations is a challenging and open research problem. To circumvent this issue, we explore two approaches to quantify explanation quality:
\begin{itemize}[noitemsep , topsep=0pt, leftmargin=*]
    \item \uline{Individual explanation accuracy}: Here, we assume there exists an oracle which has access to all the explanations, the base language guided-classifier, and the labeled examples. This oracle evaluates the accuracy of each individual explanation of the task by  evaluating it on the labeled examples with the base language-guided classifier. We term this accuracy as the individual explanation accuracy and use it a proxy for quantifying the quality of an explanation. For each of the six evaluation tasks, we provide the individual explanation accuracies in Appendix~\secref{sec:app_nlexp}. 
    
    \item \uline{Perplexity of an explanation}: Assuming access to all labeled examples (needed for the above approach) may be unrealistic for many scenarios. Hence, we also explore a surface-level metric, the perplexity of the explanation, to quantify the quality of an explanation. We obtain perplexity scores for each explanation by using the GPT2-Large pre-trained model \cite{Radford2019LanguageMA}. We provide perplexity scores of each explanation for the six evaluation tasks in Appendix~\secref{sec:app_nlexp}.
\end{itemize}

These aforementioned approaches to quantify the quality of an explanation) can filter out poor quality explanations or selectively choose good quality explanations for adapting the base language-guided classifier. We explore the following scenarios (with adaptation ratio, $\alpha = 1$) to understand the impact of explanation quality:

\begin{figure}[!h]
    \includegraphics[width=0.48\textwidth]{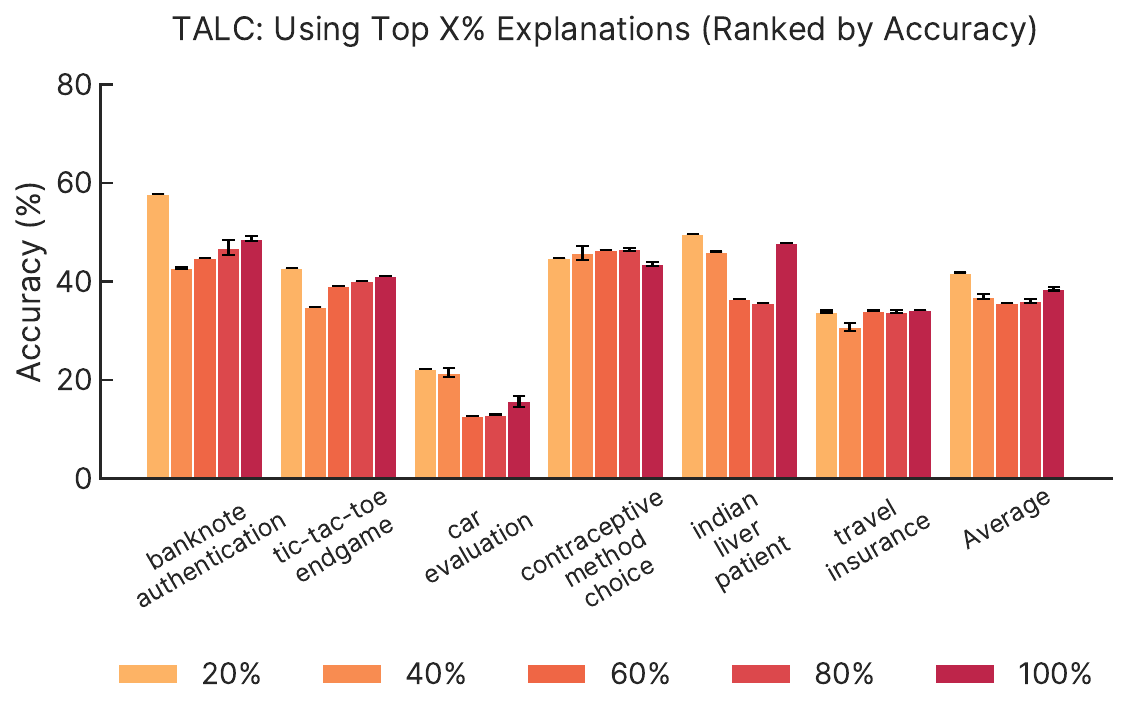}
    \par
    \includegraphics[width=0.48\textwidth]{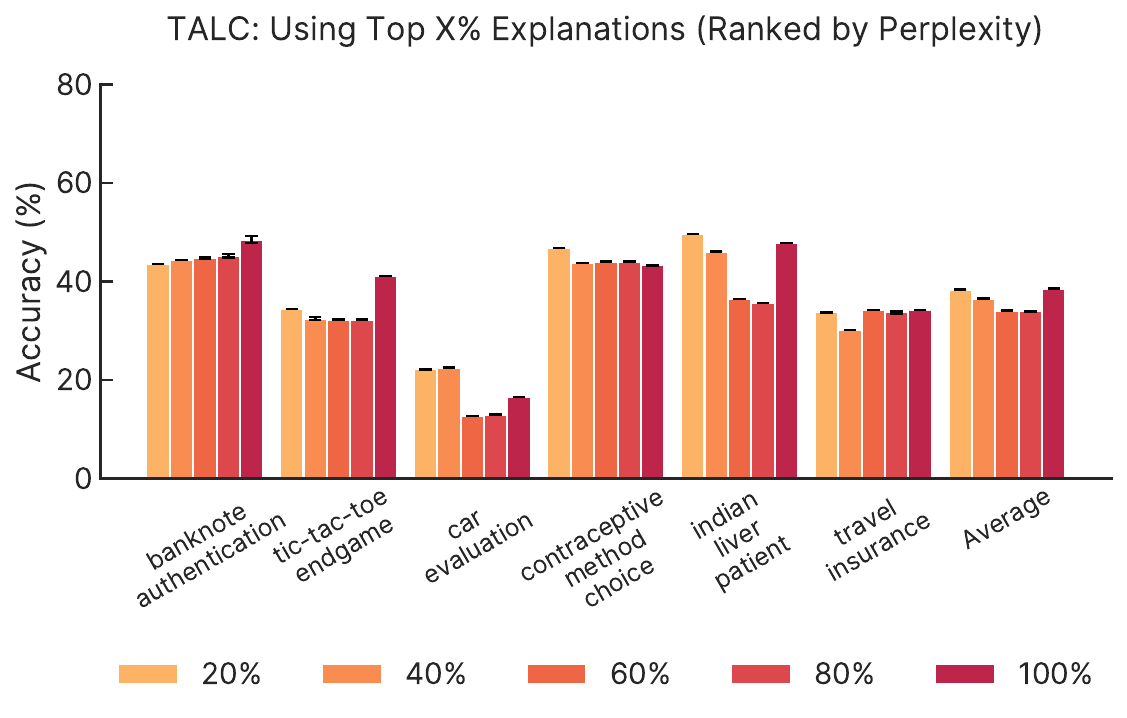}
    \vspace{-0.6cm}
    \caption{
    \nalla's performance only using the top $X\%$ explanations, where $X = 20, 40, 60, 80, 100$. On average, \nalla has the best performance when only using the explanations with the highest quality. The performance of \nalla decreases and then increases as we add explanations with lower quality. We see this trend because only the explanations with high quality are used at first and adding explanations with lower quality distract the label aggregator at first, but the label aggregator is able to distinguish high-quality explanations when the number of explanations keeps increasing. 
    }
    \label{fig:top exp}
\end{figure}
\begin{figure}[!h]
    \centering
    \includegraphics[scale=0.48]{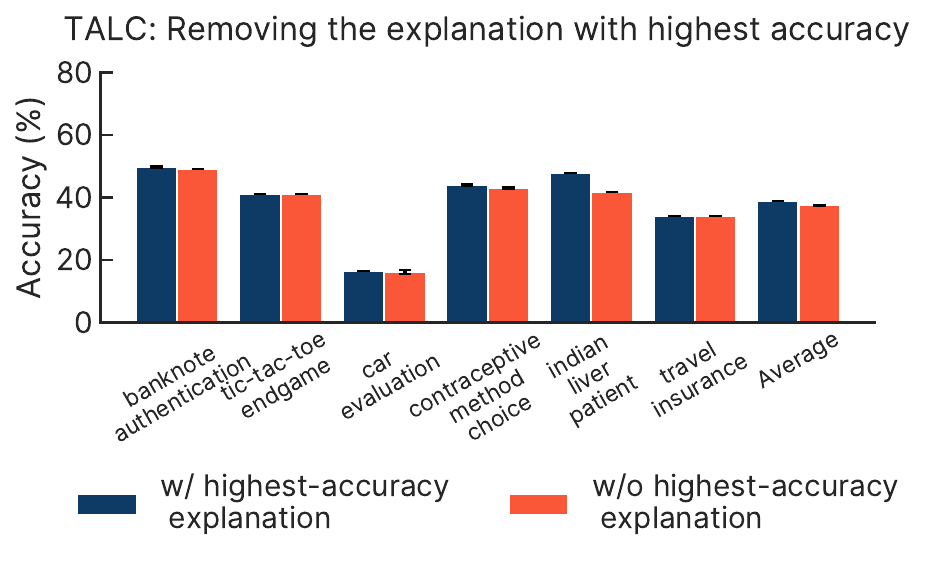}
    \caption{When ranking the explanations by their individual accuracy, removing the best explanation leads to a $1.3\%$ drop in performance on average.
    }
    \label{fig:gac no best exp}
\end{figure}
\begin{itemize}[noitemsep, topsep=0pt, leftmargin=*]
\item \underline{Using the top $X$ percentage of explanations}: 
We rank the explanations by accuracy or perplexity for each task and only use the top $X$ percent of the ranked explanations for \nalla, where $X = 20, 40, 60, 80, 100$. The results are shown in Figure~\ref{fig:top exp}. On average, we observe that \nalla performs the best when using only the top 20\% of explanations ranked by both accuracy and perplexity. As $X$ increases from $20 \rightarrow 40 \rightarrow 60$, the average performance of \nalla decreases, and then keeps increasing. We attribute this trend to the fact that the training of the label aggregator may be sub-optimal with a smaller number of explanations, and improve with more explanations.  
These results also clearly show that the label aggregator is able to distinguish explanation quality.
We note a roughly similar trend when the explanations are ranked by lowest perplexity instead of highest accuracy. This is an encouraging result, and indicates that perplexity of explanations can actually be a reasonable basis for filtering from a large pool of explanations.

\item \underline{Removing the best explanation}:
We remove the best (highest accuracy or lowest perplexity) explanation from the set of explanations for each the task and adapt \nalla. Figure~\ref{fig:gac no best exp} shows that removing the best explanation hurts performance consistently across tasks, as expected. We observe a $1.3\%$ in accuracy drop when ranking the explanations by accuracy and a $1.0\%$ drop when ranking by perplexity on average across the six tasks (shown in Appendix~\secref{sec:perplexity}). 

\begin{figure}[!h]
    \centering
    \includegraphics[scale=0.51]{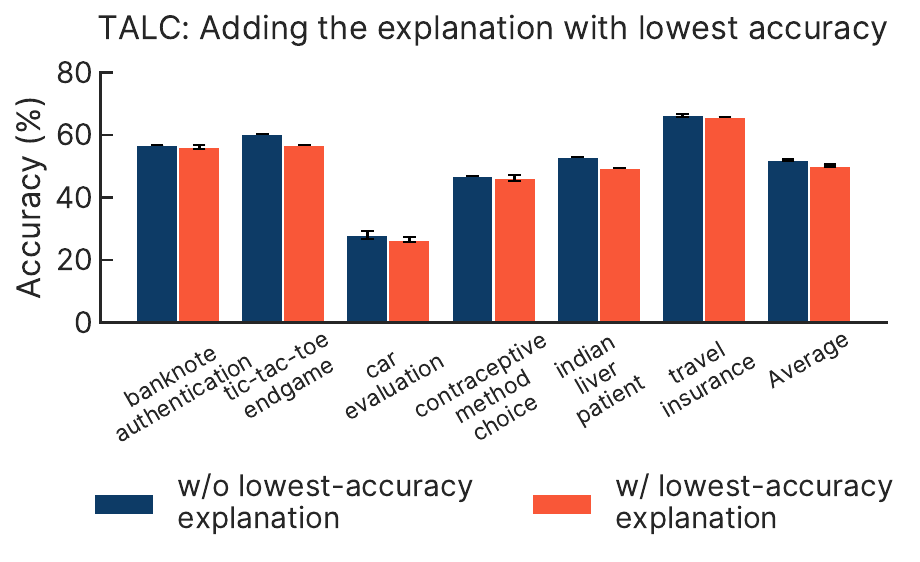}
    % \vspace{-1.5cm}
    \caption{
    Comparison of \nalla's performance before and after adding a low-quality explanation to a set of high-quality explanations. On average, the performance decreases by $1.5\%$ when ranking by accuracy.
    }
    \vspace{-0.45cm}
    \label{fig:gac one bad exp}
\end{figure}
    \item \underline{Adding a low-quality explanation to a set of} \underline{high-quality explanations}: Next, we study the impact of low-quality explanations on \nalla. For this, we consider two setups. In the first setup \nalla utilizes just the top-3 explanations as per their individual accuracies. The individual explanations accuracies can be found in Appendix~\secref{sec:app_nlexp}. 
    In the second setup, \nalla utilizes the top-3 and the worst explanation (as per individual explanation accuracy) for adaptation.
    Figure~\ref{fig:gac one bad exp} shows the performance of \nalla for these settings. When ranking by accuracy, the average decrease in performance due to the addition of low-quality explanation is $1.5\%$, demonstrating the robustness of \nalla to low-quality explanations. We observe a similar trend in results when the explanations are ranking by their perplexity (details in Appendix~\secref{sec:perplexity}).

\item \uline{Replacing best explanations with malicious explanations}: Next, we create malicious explanations by flipping the labels mentioned by the original explanations. For example, taking the explanation from Figure \ref{fig:intro} for the travel-insurance task, we convert `\textit{most college graduates have taken travel insurance}' to `\textit{most college graduates have \textbf{not} taken travel insurance}'. 
We repeat this process for the top-3 explanations ranked by accuracy or perplexity for each of the six evaluation tasks. 
\begin{figure}[!h]%
    \centering
    \includegraphics[scale=0.48]{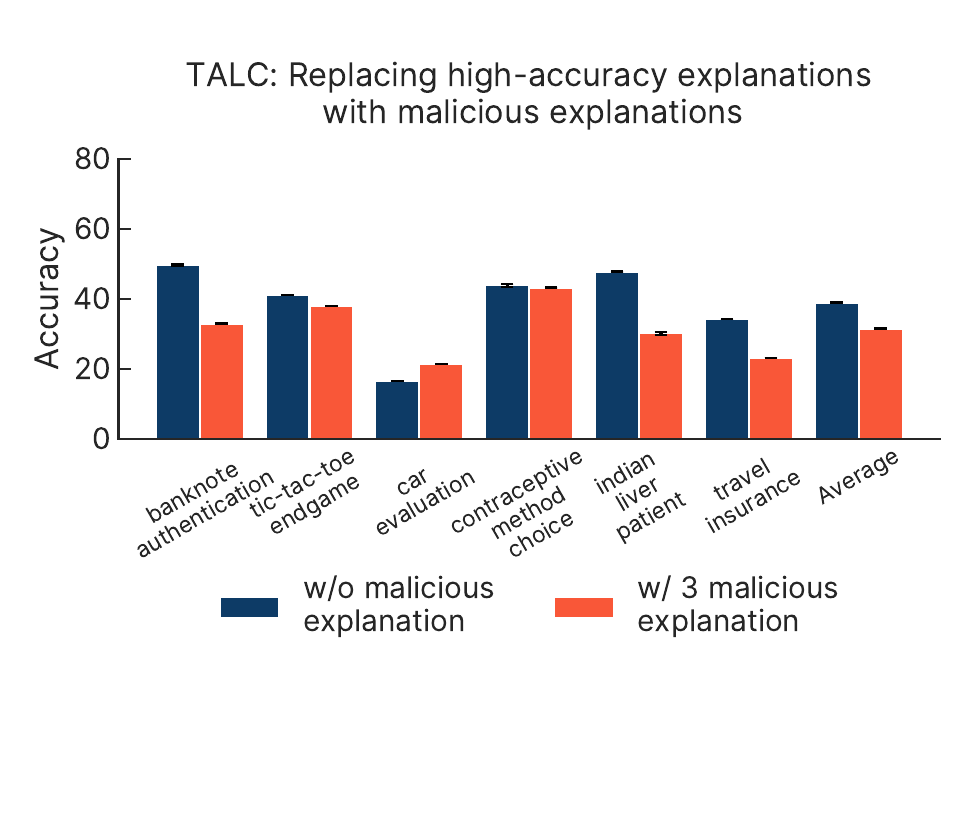} % gac_one_bad_exp
   \vspace{-2.0cm}
    \caption{
    Comparison of \nalla's performance before and after replacing good-quality explanations with malicious explanations. On average, \nalla's performance drops by $7.3\%$ as good explanations are replaced by malicious explanations.
    }
    \vspace{-0.5cm}
    \label{fig:malicious exp}
\end{figure}
The results in Figure~\ref{fig:malicious exp} show a drop in performance of \nalla (from $38.8\%$ to $31.5\%$), as expected, when the top-3 explanations (ranked by their individual accuracies) are modified into malicious explanations. When explanations are ranked by perplexity, the results are similar (details in Appendix~\secref{sec:perplexity}). 
Surprisingly, for the `car-evaluation' task, the performance increased from $16.5\%$ to $21.4\%$ on modifying the best explanations to malicious explanations when ranking by accuracy. 
From the average drop in performance, we can conclude that \nalla is susceptible to text-based attacks that may occur through the explanations provided during adaptation. Future work can address the challenge of learning to distinguish between beneficial and adversarial explanations. 
\end{itemize}

\paragraph{Agnostic nature of \nalla w.r.t language-guided classifier.}
\begin{table*}[t!]
% \normalsize
%\scriptsize
\footnotesize
\begin{center}
\begin{tabular}{ lrrr|rrr|rrr} 
 \toprule
 & \multicolumn{3}{c|}{\begin{tabular}{@{}c@{}}\textbf{T0(3B)}\end{tabular}} 
 & \multicolumn{3}{c|}{\textbf{OPT(2.7B)}} & \multicolumn{3}{c}{\textbf{FLAN-T5-XXL}} \\

\textbf{\textbf{Tasks}} & \begin{tabular}{@{}c@{}}Concat\end{tabular} & MV & \begin{tabular}{@{}c@{}}\nalla\end{tabular}
& \begin{tabular}{@{}c@{}}Concat\end{tabular} & MV & \begin{tabular}{@{}c@{}}\nalla\end{tabular} & \begin{tabular}{@{}c@{}}Concat\end{tabular} & MV & \begin{tabular}{@{}c@{}}\nalla\end{tabular}\\

 \midrule
 banknote-authentication & $44.7$ & $44.7$ & $44.7_{(0.0)}$ & $44.7$ & $45.1$ & $44.7_{(0.0)}$ & $\mathbf{54.2}$ & $31.3$ & $43.6_{(0.0)}$ \\ 
 tic-tac-toe-endgame & $\mathbf{67.7}$ & $\mathbf{67.7}$ & $\mathbf{{67.7_{(0.0)}}}$ & $32.3$ & $49.0$ & $64.9_{(1.3)}$ & $32.3$ & $40.6$ & $56.8_{(0.0)}$ \\
 car-evaluation & $3.8$ & $\mathbf{71.7}$ & $\mathbf{71.7_{(0.0)}}$ & $2.3$ & $3.8$ & $3.8_{(0.0)}$ & $3.8$ & $26.6$ & $26.6_{(0.0)}$ \\
contraceptive-choice & $23.4$ & $23.9$ & $23.4_{(0.0)}$ & $27.5$ & $33.2$ & $33.2_{(0.0)}$ & $38.3$ & $\mathbf{45.4}$ & $\mathbf{45.4_{(0.0)}}$ \\
 indian-liver-patient & $\mathbf{67.8}$ & $33.0$ & $48.7_{(0.0)}$ & $32.2$ & $32.2$ &$32.2_{(0.0)}$ & $66.1$ & $48.7$ & $56.5_{(0.4)}$ \\
 travel-insurance & $38.9$ & $65.8$ & $65.8_{(0.0)}$ & $42.5$ & $65.8$ & $65.8_{(0.0)}$ & $71.9$ & $71.9$ & $\mathbf{72.6_{(0.4)}}$ \\
 \midrule
 \textbf{Average} & $41.1$ & $51.1$ & $\mathbf{53.7}$ & $30.2$ & $38.2$ & $40.8$ & $44.4$ & $44.1$ & $50.3$ \\
\bottomrule
\end{tabular}
\end{center}
\vspace{-0.1in}
\caption{Comparison of accuracies between \nalla and baselines when using different LLMs as the language-guided classifier on the 6 different tasks from \cluesreal. We report the mean and standard deviation for the accuracy across three runs for adaptation-based methods. The numbers in \textbf{bold} indicate the best accuracy across methods.
}
\vspace{-0.1in}
\label{tab:llm-results}
\end{table*}
The flexibility of choosing different models as the underlying language-guided classifiers is an advantage of the \nalla framework. The modular design of \nalla, i.e, decoupling of (1) how we obtain predictions w.r.t each explanation using a language-guided classifier and (2) how we combine these individual predictions, makes \nalla a highly generalizable and flexible framework. 
To empirically validate this flexibility, we experiment with different LLMs as the underlying language-guided classifier. Table~\ref{tab:llm-results} compares the accuracies of \nalla and two baselines by using T0-3B \cite{sanh2022multitask}, OPT-2.7B \cite{zhang2022opt}, and Flan-T5-XXL \cite{chung2022scaling} as the underlying language-guided classifiers. As baselines, we consider two settings, (1) MV - majority vote of the predictions made by the LLM corresponding to individual explanations and (2) Concat - predicting by considering all explanations (concatenated) together in the context of the LLM. For both setting, the prediction is done by prompting the LLM. We provide the prompt templates in Appendix~\secref{sec: prompt templates}.

Table~\ref{tab:llm-results} shows that \nalla outperforms both baselines for all of the three LLMs demonstrating the robustness of \nalla to the choice of the underlying language-guided classifier.

\section{Discussion \& Conclusion}
In this paper, we introduce \nalla, a framework for test-time adaptation of language-guided classifiers that leverages multiple sources of supervision. 
One conceptual advantage of \nalla is its agnosticism towards the choice of language-guided classifier, leaving room for future exploration with different models.  \nalla is flexible in terms of what aspects of explanations, teachers and unlabeled examples are used to train the label aggregator. While our approach here trains a label-aggregator for every new task (since our features for the label aggregator include identities of individual explanations), in principle it should be possible to train a unified label aggregator across tasks based on featurized representations of tasks and explanations. Scaling up \nalla to new datasets with a larger number of tasks would provide valuable insights into its generalizability. Our experiments reveal \nalla's susceptibility to malicious attacks and bad-faith actors, which future works can improve on. Despite these challenges, 
\nalla suggests exciting opportunities for harnessing the collective wisdom of teachers in real-world applications.

\section*{Limitations}

To analyse the impact of quality of an explanation during test-time adaptation, we use individual explanation accuracy as a surrogate measure for its quality in lieu of a standardized metric of explanation quality. However, developing standardized metrics to judge the quality of an explanation remains an open and pressing research challenge.

To analyse robustness of \nalla w.r.t malicious explanations, we created malicious explanations by flipping the labels mentioned in the best explanation for a task. However, there could be other ways of creating malicious or adversarial explanations, which are more subtle than just flipping a label. For example, one subtle way of altering an existing explanation to a malicious one could be by establishing unwanted correlations between a protected attribute (e.g. gender) and the label for a downstream task (e.g. whether the loan should be approved). Analyzing and improving the robustness of \nalla to more nuanced adversarial/malicious explanations remains to be explored.

The adapted model obtained by using \nalla is task dependent, as it uses explanations and unlabeled data specific to the downstream task for adaptation (specifically, for training the label aggregator component). Hence, for every novel task for which we want a adapt a base language-guided classifier, we need access to explanations and unlabeled samples. This requirement (especially obtaining good explanations for adaptation) can be a challenging issue for some real-world scenarios. Improving \nalla to reduce its dependence on the amount of explanations and/or unlabeled data while still retaining downstream accuracy (post-adaptation) is an interesting direction for future work. The base language-guided classifier used in our experiments, \exent, is designed to work with a maximum of 512 tokens in its context. Usage of longer context models or even large-scale pre-trained models remains to be explored.
The effectiveness of \nalla under multilingual setting is also unexplored. 

\section*{Ethics and Broader Impact}
The experiments described in this work are performed over tasks from a publicly available benchmark, \clues~\cite{menon-etal-2022-clues}. The data for these tasks do not contain any personally identifiable information. We do not collect or annotate any additional data. For all the experiments in the paper we evaluate using automatic metrics and do not perform any human evaluation.

\nalla is agnostic to the base language-guided classifier. We do not foresee major risks with our framework if the inputs provided are appropriate. Like any other natural language guided method there are potential concerns of misguiding a model deliberately by providing erroneous inputs. Measures to detect such bad actors and rectifying erroneous inputs is beyond the scope of this work. However, there is a risk of classifiers perpetuating biases present in the input natural language explanations (for example, some explanations may describe the label in terms of sensitive or inappropriate features). Biased or discriminatory explanations can result in biased predictions and contribute to unjust outcomes. 

The broader impact of this work can lead to development of frameworks that enable efficient adaptation of AI systems. 
Developing language-guided adaptable systems can improve the impact and usability of AI systems in daily life, especially on the long tail of tasks with limited labeled data. However, the responsible development and deployment of these models would require domain-specific expertise, involving collaboration with experts and stakeholders to understand the implications and ensure ethical considerations are met. Close attention should be paid to the specific contexts in which the classifiers are applied to minimize negative consequences and maximize positive impacts.

\section*{Acknowledgments}

The authors would like to thank the anonymous reviewers for their suggestions and feedback on the work. This work was supported in part by NSF grant DRL2112635. The views contained in this article are those of the authors and not of the funding agency.

\bibliography{anthology,custom}

\begin{thebibliography}{30}
\expandafter\ifx\csname natexlab\endcsname\relax\def\natexlab#1{#1}\fi

\bibitem[{Andreas et~al.(2018)Andreas, Klein, and
  Levine}]{andreas-etal-2018-learning}
Jacob Andreas, Dan Klein, and Sergey Levine. 2018.
\newblock \href {https://doi.org/10.18653/v1/N18-1197} {Learning with latent
  language}.
\newblock In \emph{Proceedings of the 2018 Conference of the North {A}merican
  Chapter of the Association for Computational Linguistics: Human Language
  Technologies, Volume 1 (Long Papers)}, pages 2166--2179, New Orleans,
  Louisiana. Association for Computational Linguistics.

\bibitem[{Banerjee et~al.(2021)Banerjee, Gokhale, and
  Baral}]{banerjee-etal-2021-self}
Pratyay Banerjee, Tejas Gokhale, and Chitta Baral. 2021.
\newblock \href {https://doi.org/10.18653/v1/2021.naacl-main.95}
  {Self-supervised test-time learning for reading comprehension}.
\newblock In \emph{Proceedings of the 2021 Conference of the North American
  Chapter of the Association for Computational Linguistics: Human Language
  Technologies}, pages 1200--1211, Online. Association for Computational
  Linguistics.

\bibitem[{Chopra et~al.(2019)Chopra, Tessler, and Goodman}]{chopra2019first}
Sahil Chopra, Michael~Henry Tessler, and Noah~D Goodman. 2019.
\newblock The first crank of the cultural ratchet: Learning and transmitting
  concepts through language.
\newblock In \emph{Proceedings of the 41st Annual Meeting of the Cognitive
  Science Society}, pages 226--232.

\bibitem[{Chung et~al.(2022)Chung, Hou, Longpre, Zoph, Tay, Fedus, Li, Wang,
  Dehghani, Brahma, Webson, Gu, Dai, Suzgun, Chen, Chowdhery, Castro-Ros,
  Pellat, Robinson, Valter, Narang, Mishra, Yu, Zhao, Huang, Dai, Yu, Petrov,
  Chi, Dean, Devlin, Roberts, Zhou, Le, and Wei}]{chung2022scaling}
Hyung~Won Chung, Le~Hou, Shayne Longpre, Barret Zoph, Yi~Tay, William Fedus,
  Yunxuan Li, Xuezhi Wang, Mostafa Dehghani, Siddhartha Brahma, Albert Webson,
  Shixiang~Shane Gu, Zhuyun Dai, Mirac Suzgun, Xinyun Chen, Aakanksha
  Chowdhery, Alex Castro-Ros, Marie Pellat, Kevin Robinson, Dasha Valter,
  Sharan Narang, Gaurav Mishra, Adams Yu, Vincent Zhao, Yanping Huang, Andrew
  Dai, Hongkun Yu, Slav Petrov, Ed~H. Chi, Jeff Dean, Jacob Devlin, Adam
  Roberts, Denny Zhou, Quoc~V. Le, and Jason Wei. 2022.
\newblock \href {http://arxiv.org/abs/2210.11416} {Scaling
  instruction-finetuned language models}.

\bibitem[{Devlin et~al.(2019)Devlin, Chang, Lee, and
  Toutanova}]{devlin-etal-2019-bert}
Jacob Devlin, Ming-Wei Chang, Kenton Lee, and Kristina Toutanova. 2019.
\newblock \href {https://doi.org/10.18653/v1/N19-1423} {{BERT}: Pre-training of
  deep bidirectional transformers for language understanding}.
\newblock In \emph{Proceedings of the 2019 Conference of the North {A}merican
  Chapter of the Association for Computational Linguistics: Human Language
  Technologies, Volume 1 (Long and Short Papers)}, pages 4171--4186,
  Minneapolis, Minnesota. Association for Computational Linguistics.

\bibitem[{Goldwasser and Roth(2014)}]{7524b90aadb74d5abcd045aaf8407779}
Dan Goldwasser and Dan Roth. 2014.
\newblock \href {https://doi.org/10.1007/s10994-013-5407-y} {Learning from
  natural instructions}.
\newblock \emph{Machine Learning}, 94(2):205--232.

\bibitem[{Gonen et~al.(2022)Gonen, Iyer, Blevins, Smith, and
  Zettlemoyer}]{gonen2022demystifying}
Hila Gonen, Srini Iyer, Terra Blevins, Noah~A. Smith, and Luke Zettlemoyer.
  2022.
\newblock \href {http://arxiv.org/abs/2212.04037} {Demystifying prompts in
  language models via perplexity estimation}.

\bibitem[{Hancock et~al.(2018)Hancock, Varma, Wang, Bringmann, Liang, and
  R{\'e}}]{hancock-etal-2018-training}
Braden Hancock, Paroma Varma, Stephanie Wang, Martin Bringmann, Percy Liang,
  and Christopher R{\'e}. 2018.
\newblock \href {https://doi.org/10.18653/v1/P18-1175} {Training classifiers
  with natural language explanations}.
\newblock In \emph{Proceedings of the 56th Annual Meeting of the Association
  for Computational Linguistics (Volume 1: Long Papers)}, pages 1884--1895,
  Melbourne, Australia. Association for Computational Linguistics.

\bibitem[{Iwasawa and Matsuo(2021)}]{NEURIPS2021_1415fe9f}
Yusuke Iwasawa and Yutaka Matsuo. 2021.
\newblock \href
  {https://proceedings.neurips.cc/paper/2021/file/1415fe9fea0fa1e45dddcff5682239a0-Paper.pdf}
  {Test-time classifier adjustment module for model-agnostic domain
  generalization}.
\newblock In \emph{Advances in Neural Information Processing Systems},
  volume~34, pages 2427--2440. Curran Associates, Inc.

\bibitem[{Khurana et~al.(2021)Khurana, Paul, Rai, Biswas, and
  Aggarwal}]{https://doi.org/10.48550/arxiv.2112.02355}
Ansh Khurana, Sujoy Paul, Piyush Rai, Soma Biswas, and Gaurav Aggarwal. 2021.
\newblock \href {https://doi.org/10.48550/ARXIV.2112.02355} {Sita: Single image
  test-time adaptation}.

\bibitem[{Lu et~al.(2022)Lu, Bartolo, Moore, Riedel, and
  Stenetorp}]{lu2022fantastically}
Yao Lu, Max Bartolo, Alastair Moore, Sebastian Riedel, and Pontus Stenetorp.
  2022.
\newblock \href {http://arxiv.org/abs/2104.08786} {Fantastically ordered
  prompts and where to find them: Overcoming few-shot prompt order
  sensitivity}.

\bibitem[{Murty et~al.(2020)Murty, Koh, and Liang}]{murty-etal-2020-expbert}
Shikhar Murty, Pang~Wei Koh, and Percy Liang. 2020.
\newblock \href {https://doi.org/10.18653/v1/2020.acl-main.190} {{E}xp{BERT}:
  Representation engineering with natural language explanations}.
\newblock In \emph{Proceedings of the 58th Annual Meeting of the Association
  for Computational Linguistics}, pages 2106--2113, Online. Association for
  Computational Linguistics.

\bibitem[{Nado et~al.(2020)Nado, Padhy, Sculley, D'Amour, Lakshminarayanan, and
  Snoek}]{https://doi.org/10.48550/arxiv.2006.10963}
Zachary Nado, Shreyas Padhy, D.~Sculley, Alexander D'Amour, Balaji
  Lakshminarayanan, and Jasper Snoek. 2020.
\newblock \href {https://doi.org/10.48550/ARXIV.2006.10963} {Evaluating
  prediction-time batch normalization for robustness under covariate shift}.

\bibitem[{R.~Menon et~al.(2022)R.~Menon, Ghosh, and
  Srivastava}]{menon-etal-2022-clues}
Rakesh R.~Menon, Sayan Ghosh, and Shashank Srivastava. 2022.
\newblock \href {https://doi.org/10.18653/v1/2022.acl-long.451} {{CLUES}: A
  benchmark for learning classifiers using natural language explanations}.
\newblock In \emph{Proceedings of the 60th Annual Meeting of the Association
  for Computational Linguistics (Volume 1: Long Papers)}, pages 6523--6546,
  Dublin, Ireland. Association for Computational Linguistics.

\bibitem[{Radford et~al.(2019)Radford, Wu, Child, Luan, Amodei, Sutskever
  et~al.}]{Radford2019LanguageMA}
Alec Radford, Jeffrey Wu, Rewon Child, David Luan, Dario Amodei, Ilya
  Sutskever, et~al. 2019.
\newblock Language models are unsupervised multitask learners.
\newblock \emph{OpenAI blog}, 1(8):9.

\bibitem[{Ratner et~al.(2018{\natexlab{a}})Ratner, Hancock, Dunnmon, Goldman,
  and R{\'e}}]{ratner2018snorkel}
Alex Ratner, Braden Hancock, Jared Dunnmon, Roger Goldman, and Christopher
  R{\'e}. 2018{\natexlab{a}}.
\newblock Snorkel metal: Weak supervision for multi-task learning.
\newblock In \emph{Proceedings of the Second Workshop on Data Management for
  End-To-End Machine Learning}, pages 1--4.

\bibitem[{Ratner et~al.(2017)Ratner, Bach, Ehrenberg, Fries, Wu, and R{\'{e}
  }}]{Ratner_2017}
Alexander Ratner, Stephen~H. Bach, Henry Ehrenberg, Jason Fries, Sen Wu, and
  Christopher R{\'{e} }. 2017.
\newblock \href {https://doi.org/10.14778/3157794.3157797} {Snorkel}.
\newblock \emph{Proceedings of the {VLDB} Endowment}, 11(3):269--282.

\bibitem[{Ratner et~al.(2018{\natexlab{b}})Ratner, Hancock, Dunnmon, Goldman,
  and R{\'{e} }}]{Ratner_2018}
Alexander Ratner, Braden Hancock, Jared Dunnmon, Roger Goldman, and Christopher
  R{\'{e} }. 2018{\natexlab{b}}.
\newblock \href {https://doi.org/10.1145/3209889.3209898} {Snorkel metal: Weak
  supervision for multi-task learning}.
\newblock \emph{In DEEM’18: International Workshop on Data Management for
  End-to-End Machine Learning}, 11(3):4.

\bibitem[{Ratner et~al.(2019)Ratner, Hancock, Dunnmon, Sala, Pandey, and
  R{\'e}}]{ratner2019training}
Alexander Ratner, Braden Hancock, Jared Dunnmon, Frederic Sala, Shreyash
  Pandey, and Christopher R{\'e}. 2019.
\newblock Training complex models with multi-task weak supervision.
\newblock In \emph{Proceedings of the AAAI Conference on Artificial
  Intelligence}, volume~33, pages 4763--4771.

\bibitem[{Sanh et~al.(2022)Sanh, Webson, Raffel, Bach, Sutawika, Alyafeai,
  Chaffin, Stiegler, Raja, Dey, Bari, Xu, Thakker, Sharma, Szczechla, Kim,
  Chhablani, Nayak, Datta, Chang, Jiang, Wang, Manica, Shen, Yong, Pandey,
  Bawden, Wang, Neeraj, Rozen, Sharma, Santilli, Fevry, Fries, Teehan, Scao,
  Biderman, Gao, Wolf, and Rush}]{sanh2022multitask}
Victor Sanh, Albert Webson, Colin Raffel, Stephen Bach, Lintang Sutawika, Zaid
  Alyafeai, Antoine Chaffin, Arnaud Stiegler, Arun Raja, Manan Dey, M~Saiful
  Bari, Canwen Xu, Urmish Thakker, Shanya~Sharma Sharma, Eliza Szczechla,
  Taewoon Kim, Gunjan Chhablani, Nihal Nayak, Debajyoti Datta, Jonathan Chang,
  Mike Tian-Jian Jiang, Han Wang, Matteo Manica, Sheng Shen, Zheng~Xin Yong,
  Harshit Pandey, Rachel Bawden, Thomas Wang, Trishala Neeraj, Jos Rozen,
  Abheesht Sharma, Andrea Santilli, Thibault Fevry, Jason~Alan Fries, Ryan
  Teehan, Teven~Le Scao, Stella Biderman, Leo Gao, Thomas Wolf, and Alexander~M
  Rush. 2022.
\newblock \href {https://openreview.net/forum?id=9Vrb9D0WI4} {Multitask
  prompted training enables zero-shot task generalization}.
\newblock In \emph{International Conference on Learning Representations}.

\bibitem[{Schneider et~al.(2020)Schneider, Rusak, Eck, Bringmann, Brendel, and
  Bethge}]{https://doi.org/10.48550/arxiv.2006.16971}
Steffen Schneider, Evgenia Rusak, Luisa Eck, Oliver Bringmann, Wieland Brendel,
  and Matthias Bethge. 2020.
\newblock \href {https://doi.org/10.48550/ARXIV.2006.16971} {Improving
  robustness against common corruptions by covariate shift adaptation}.

\bibitem[{Sivaprasad and
  Fleuret(2021)}]{https://doi.org/10.48550/arxiv.2110.10232}
Prabhu~Teja Sivaprasad and François Fleuret. 2021.
\newblock \href {https://doi.org/10.48550/ARXIV.2110.10232} {Test time
  adaptation through perturbation robustness}.

\bibitem[{Srivastava et~al.(2017)Srivastava, Labutov, and
  Mitchell}]{srivastava-etal-2017-joint}
Shashank Srivastava, Igor Labutov, and Tom Mitchell. 2017.
\newblock \href {https://doi.org/10.18653/v1/D17-1161} {Joint concept learning
  and semantic parsing from natural language explanations}.
\newblock In \emph{Proceedings of the 2017 Conference on Empirical Methods in
  Natural Language Processing}, pages 1527--1536, Copenhagen, Denmark.
  Association for Computational Linguistics.

\bibitem[{Sun et~al.(2020)Sun, Wang, Zhuang, Miller, Hardt, and
  Efros}]{sun19ttt}
Yu~Sun, Xiaolong Wang, Liu Zhuang, John Miller, Moritz Hardt, and Alexei~A.
  Efros. 2020.
\newblock Test-time training with self-supervision for generalization under
  distribution shifts.
\newblock In \emph{ICML}.

\bibitem[{Tomasello(1999)}]{10.2307/j.ctvjsf4jc}
Michael Tomasello. 1999.
\newblock \href {http://www.jstor.org/stable/j.ctvjsf4jc} {\emph{The Cultural
  Origins of Human Cognition}}.
\newblock Harvard University Press.

\bibitem[{Wang et~al.(2020)Wang, Shelhamer, Liu, Olshausen, and
  Darrell}]{https://doi.org/10.48550/arxiv.2006.10726}
Dequan Wang, Evan Shelhamer, Shaoteng Liu, Bruno Olshausen, and Trevor Darrell.
  2020.
\newblock \href {https://doi.org/10.48550/ARXIV.2006.10726} {Tent: Fully
  test-time adaptation by entropy minimization}.

\bibitem[{Wang* et~al.(2020)Wang*, Qin*, Zhou, Yan, Ye, Neves, Liu, and
  Ren}]{Wang*2020Learning}
Ziqi Wang*, Yujia Qin*, Wenxuan Zhou, Jun Yan, Qinyuan Ye, Leonardo Neves,
  Zhiyuan Liu, and Xiang Ren. 2020.
\newblock \href {https://openreview.net/forum?id=rJlUt0EYwS} {Learning from
  explanations with neural execution tree}.
\newblock In \emph{International Conference on Learning Representations}.

\bibitem[{Ye et~al.(2020)Ye, Huang, and
  Ren}]{DBLP:journals/corr/abs-2005-00806}
Qinyuan Ye, Xiao Huang, and Xiang Ren. 2020.
\newblock \href {http://arxiv.org/abs/2005.00806} {Teaching machine
  comprehension with compositional explanations}.
\newblock \emph{CoRR}, abs/2005.00806.

\bibitem[{Zhang et~al.(2021)Zhang, Levine, and
  Finn}]{https://doi.org/10.48550/arxiv.2110.09506}
Marvin Zhang, Sergey Levine, and Chelsea Finn. 2021.
\newblock \href {https://doi.org/10.48550/ARXIV.2110.09506} {Memo: Test time
  robustness via adaptation and augmentation}.

\bibitem[{Zhang et~al.(2022)Zhang, Roller, Goyal, Artetxe, Chen, Chen, Dewan,
  Diab, Li, Lin, Mihaylov, Ott, Shleifer, Shuster, Simig, Koura, Sridhar, Wang,
  and Zettlemoyer}]{zhang2022opt}
Susan Zhang, Stephen Roller, Naman Goyal, Mikel Artetxe, Moya Chen, Shuohui
  Chen, Christopher Dewan, Mona Diab, Xian Li, Xi~Victoria Lin, Todor Mihaylov,
  Myle Ott, Sam Shleifer, Kurt Shuster, Daniel Simig, Punit~Singh Koura, Anjali
  Sridhar, Tianlu Wang, and Luke Zettlemoyer. 2022.
\newblock \href {http://arxiv.org/abs/2205.01068} {Opt: Open pre-trained
  transformer language models}.

\end{thebibliography}
\bibliographystyle{acl_natbib}

\newpage
\section*{Appendix} \label{sec:appendix}
\addcontentsline{toc}{subsection}{Appendix}
\renewcommand{\thesubsection}{\Alph{subsection}}

\subsection{Usage of \nalla} \label{sec:talc_usage}
\begin{enumerate}
   \item An example of real-world cases where the entire set of test samples can be realistically accessed:
   
   Let's consider the case of a product category classifier for products on the Amazon database. In this case, developers will first define classifiers using some training data and deploy the classifier on the entire database to label examples.

   \item Method to use TALC when test samples are observed one-by-one:
   
   Even if we do not have access to the entire test set and the classifier observes unlabeled samples one by one, TALC can be deployed in practice as:

   \begin{enumerate}
     \item For a predetermined number of samples, the language-guided classifier is deployed off-the-shelf (Note: In this work, this would be the same as using \exent for those samples).
     \item The aforementioned samples can now be pooled together as an adaptation set, and we can adapt the language-guided classifier using \nalla.
   \end{enumerate}
   In other words, we incur a “warm-up”  phase, where the un-adapted classifier is used, following which we adapt the classifier using TALC by considering the set of samples observed during warm-up as an adaptation set.
 \end{enumerate}

\subsection{Details of evaluation tasks} \label{sec:app_data}
We use 6 real world classification tasks from \citet{menon-etal-2022-clues} as our evaluation tasks. 
The tasks considered are -- uci/banknote-authentication, uci/tic-tac-toe-endgame, uci/car-evaluation,  uci/contraceptive-method-choice, uci/indian-liver-patient, and kaggle/travel-insurance. 
Examples of these tasks can be found at the \clues website with the following link: \url{https://clues-benchmark.github.io}. 
Among the above tasks,  uci/car-evaluation and  uci/contraceptive-method-choice are multi-class classification tasks while the rest tasks are binary classification tasks. 
The numbers of examples in test set of each task are 275, 195, 346, 295, 115, 398 for uci/banknote-authentication, uci/tic-tac-toe-endgame, uci/car-evaluation,  uci/contraceptive-method-choice, uci/indian-liver-patient, and kaggle/travel-insurance respectively. 

\subsection{Hyperparameter and Compute Details} \label{sec:app_hyp}
We train the \exent model following the hyperparameters in \citet{menon-etal-2022-clues}, e.g. a learning rate of 1e-5 for 5 epochs, batch size of 2, and evaluation batch size of 16. For the label aggregator training, we did hyper-parameter search for each of the tasks and report the best hyper-parameters in Table~\ref{tab:label aggregator hyperparameter}.

For fine-tuning the \exent model, compute time ranged from 1 hr for the shortest jobs with smaller data sizes to 2 hours on 1 RTX 2080Ti GPU. For fine-tuning the label aggregator, compute time is within 1 minute.

\subsection{Prompt Templates for LLM Experiments}
\label{sec: prompt templates}
For the experiments using large language models, we used the prompts elaborated in Table \ref{tab:prompts}.

\begin{table}[]
    \centering
    \begin{tabular}{p{1cm}|p{5.5cm}}
    \hline
        \textbf{Models} & \textbf{Prompt} \\
        \hline
        
        \begin{tabular}{p{1cm}}T0 \\ {\small FLAN-T5}\end{tabular} & 
        \begin{tabular}{p{5.5cm}}
        {\small
             Explanations: <explanation\_1>. <explanation\_2>. \dots. <explanation\_n> Note Details: <feat\_1> equal to <feat\_1\_value>. <feat\_2> equal to <feat\_2\_value> \dots From the explanations, is the note fake or original? Answer: }
        \end{tabular}
        \\
        \hline
        OPT & 
        \begin{tabular}{p{5.5cm}}
        {\small
             The following is a classification task that uses the following explanations. Based on the explanation classify the subsequent sample:} \\
             \\
             {\small Explanations:}\\
             {\small - <explanation\_1>}\\
             {\small - <explanation\_2>}\\
             \vdots
             {\small - <explanation\_n>}\\
             \\
             {\small Note Details: <feat\_1> equal to <feat\_1\_value>. <feat\_2> equal to <feat\_2\_value> \dots}\\
             {\small From the explanations, is the note fake or original?}\\
             \\
             {\small Answer:}
        \end{tabular}\\
    \hline
    \end{tabular}
    \caption{Prompt templates used for large language model experiments in \ref{sec:analysis}.}
    \label{tab:prompts}
\end{table}

\begin{table}[h!]
% \centering
\scalebox{0.8}{
 \begin{tabular}{l|r|r} 
 \hline
 \Thead{Task} & \Thead{Learning Rate} & \Thead{Epoch} \\
 \hline
 banknote-authentication & $4.38e-08$ & $500$\\
 \hline
 tic-tac-toe-endgame & $7.31e-08$ & $40$\\
 \hline
 car-evaluation & $2.48e-04$  & $500$\\
 \hline
 contraceptive-method-choice & $5.53e-03$ & $100$\\
 \hline
 indian-liver-patient &$7.91e-03$  & $50$\\
 \hline
 travel-insurance & $7.31e-08$ & $40$\\
 \hline
 \end{tabular}}
 \caption{The best hyper-parameters for training label aggregator.}
 \label{tab:label aggregator hyperparameter}
\end{table}

\subsection{Learned Label Aggregator Explanation Weight} \label{sec:app_la_weight}
\begin{figure*}[!h]
    \centering
    \begin{subfigure}[b]{0.45\textwidth}
    \centering
    \includegraphics[scale=0.45]{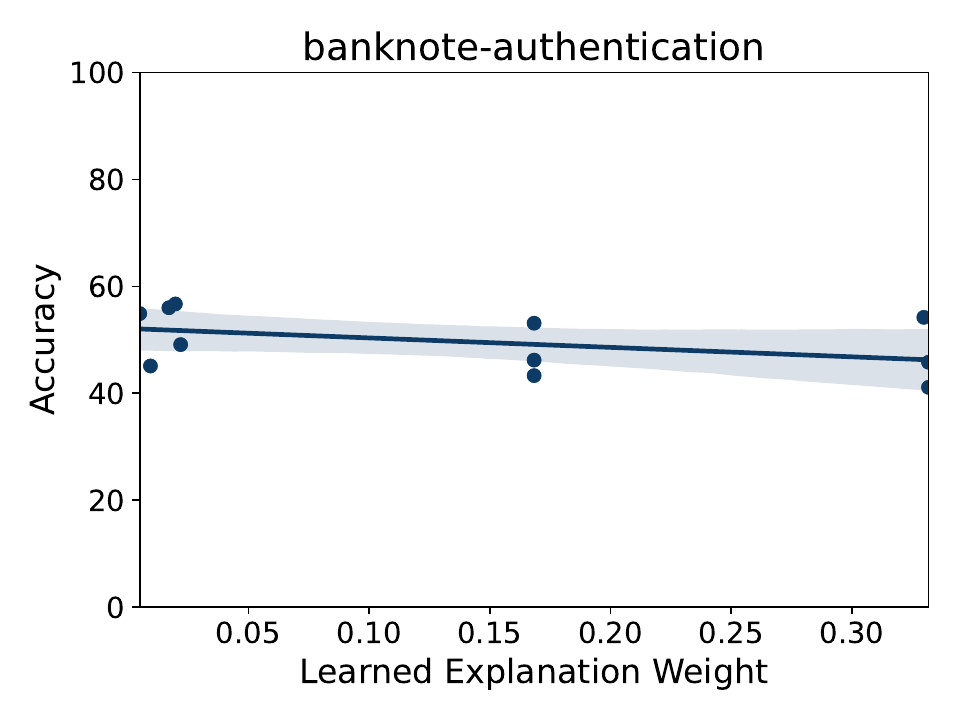}
   \vspace{-0.3cm}
    \caption{Pearson correlation coefficient = -0.12.}
    \label{fig:banknote-authentication_accuracy_factor_trend}
    \end{subfigure}
    \hfill
    \begin{subfigure}[b]{0.45\textwidth}
    \centering
    \includegraphics[scale=0.45]{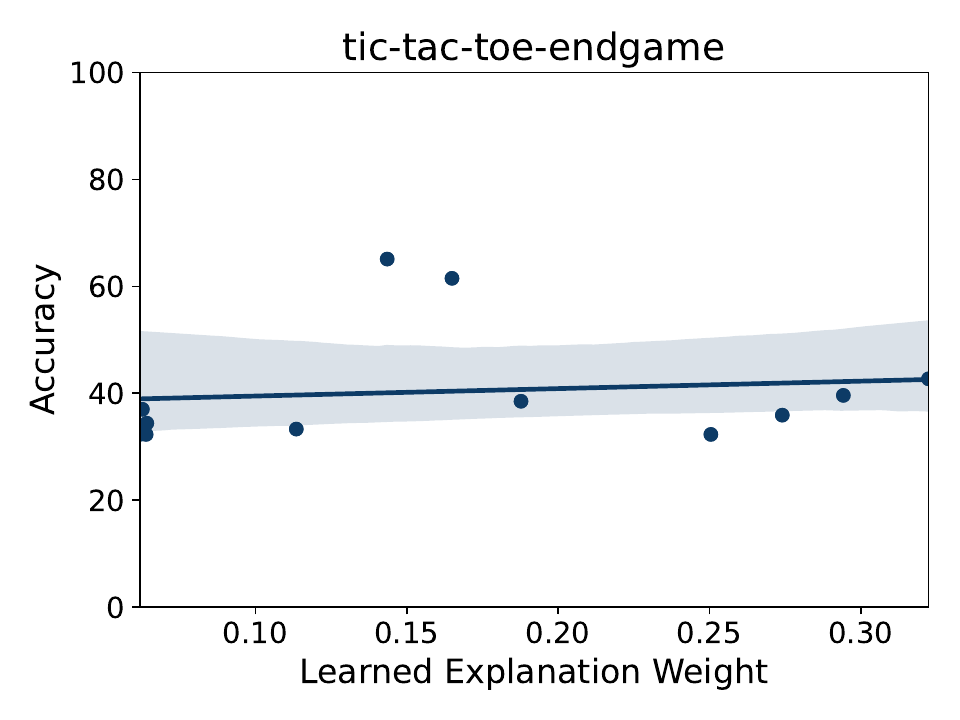}
   \vspace{-0.3cm}
    \caption{Pearson correlation coefficient = 0.12.}
    \label{fig:car_evaluation_accuracy_factor_trend}
    \end{subfigure}
    \hfill
    \begin{subfigure}[b]{0.45\textwidth}
    \centering
    \includegraphics[scale=0.45]{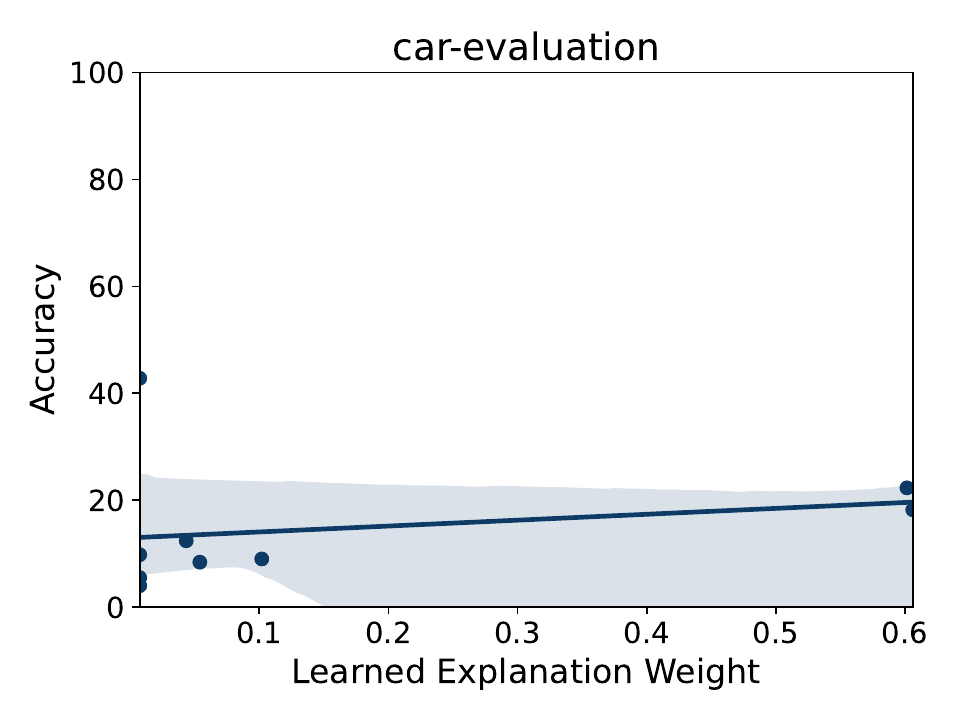}
   \vspace{-0.3cm}
    \caption{Pearson correlation coefficient = 0.23.}
    \label{fig:car_evaluation_accuracy_factor_trend}
    \end{subfigure}
    \hfill
    \begin{subfigure}[b]{0.45\textwidth}
        \centering
        \includegraphics[scale=0.45]{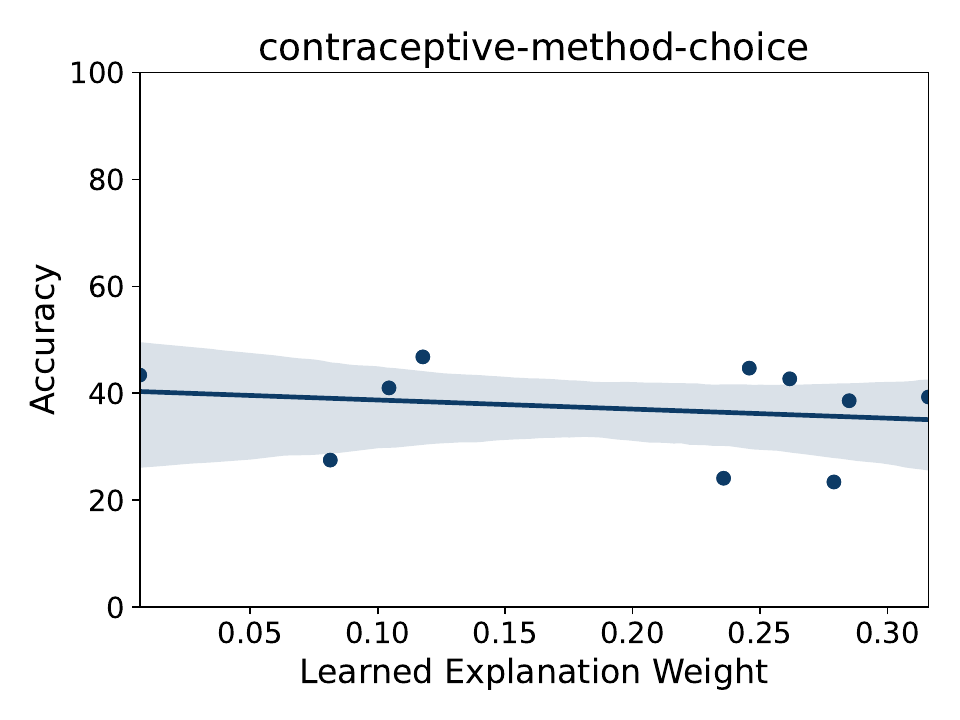}
       \vspace{-0.3cm}
        \caption{Pearson correlation coefficient = -0.21.}
        \label{fig:contraceptive-choice_accuracy_factor_trend}
    \end{subfigure}
    \hfill
    \begin{subfigure}[b]{0.45\textwidth}
        \centering
        \includegraphics[scale=0.45]{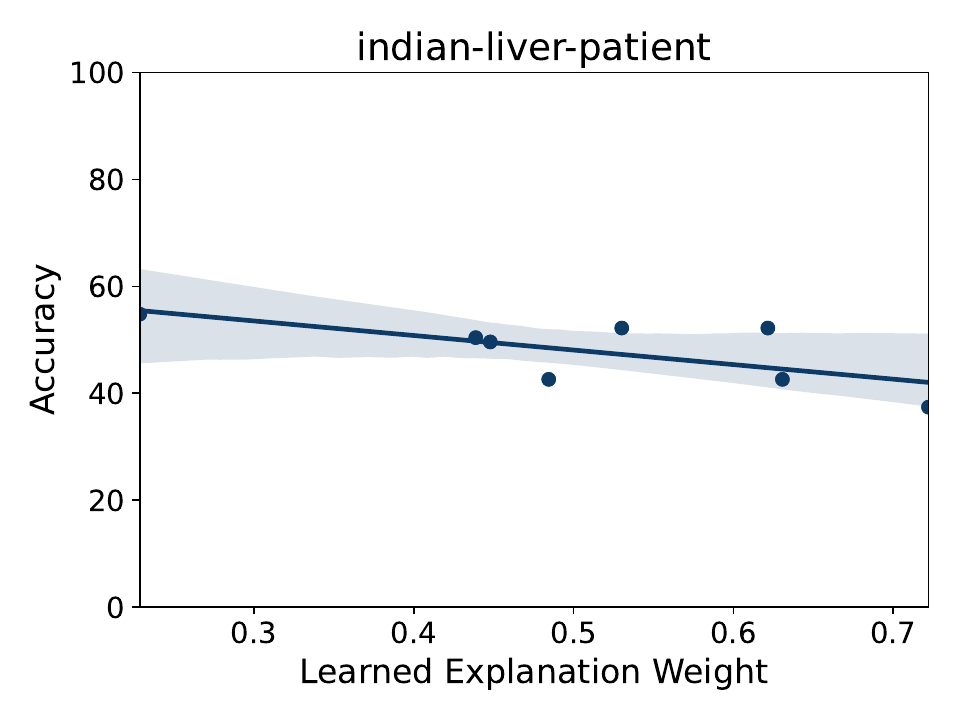}
       \vspace{-0.3cm}
        \caption{Pearson correlation coefficient = -0.68.}
        \label{fig:indian-liver-patient_accuracy_factor_trend}
    \end{subfigure}
    \hfill
    \begin{subfigure}[b]{0.45\textwidth}
        \centering
        \includegraphics[scale=0.45]{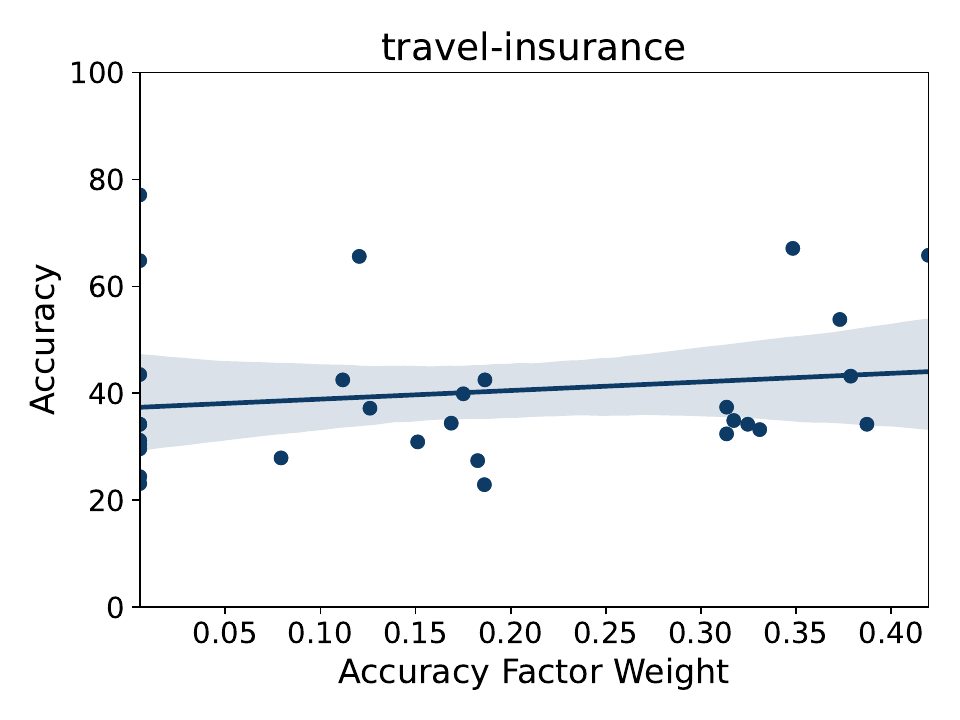}
       \vspace{-0.3cm}
        \caption{Pearson correlation coefficient = 0.16.}
        \label{fig:travel-insurance_accuracy_factor_trend}
    \end{subfigure}
    \caption{Relationship between learned accuracy factor weight and the corresponding explanations' accuracy for (a) banknote-authentication, (b) tic-tac-toe-endgame, (c) car-evaluation, (d) contraceptive-method-choice, (e) indian-liver-patient, and (f) travel-insurance.}
    \label{fig:accuracy-weight}
\end{figure*}

We analyze the learned values of the weights in the label aggregator, $\mathcal{L}^{\textrm{agg}}_w$,
to interpret the contribution of each explanation towards the final prediction of \nalla at an adaptation ratio of 1.0. First, we calculate the accuracy of each individual explanation of a task by using it with \exent for classification on the entire test set. These individual explanation accuracies serve as a proxy for their relative quality. The visualization for the 6 datasets' learned explanation weights of label aggregators and the learned explanation weight trends are shown in Figure~\ref{fig:accuracy-weight}.

Here, we also show the average learned explanation weight for explanations with/without quantifiers and the average learned explanation weight for explanations with/without conjunctions in Table~\ref{tab:quantifier-conjunction_learned_accuracy_factor weight}. Quantifiers are words like 'always' and 'usually'. Conjunctions are words like 'and' and 'or'. We use the same quantifier and conjunction words following \citet{menon-etal-2022-clues}

\begin{table}[h!]
% \small
\begin{center}
\scalebox{0.58}{
\begin{tabular}{ crr|rr} 
 \toprule
\textbf{\textbf{Dataset}} & \textbf{Quantifier} & \begin{tabular}{@{}c@{}}\textbf{No Quantifier}\end{tabular} 
& \begin{tabular}{@{}c@{}}\textbf{Conjunction}\end{tabular} & \textbf{No Conjunction} \\
 \midrule
 \begin{tabular}{@{}c@{}}banknote\\authentication\end{tabular}  & $0.18$ & $0.14$  & - & $0.14$ \\
 % \hline
 \begin{tabular}{@{}c@{}}tic-tac-toe\\endgame\end{tabular} & $0.20$ & $0.12$  & $0.29$ & $0.16$  \\ 
   % \hline
 \begin{tabular}{@{}c@{}}car\\evaluation\end{tabular} & $0.01$  & $0.18$  & $0.15$ & $0.18$ \\ 
  % \hline
\begin{tabular}{@{}c@{}}contraceptive\\choice\end{tabular} & $0.15$ & $0.21$  & $0.14$ & $0.22$ \\ 
 % \hline
 \begin{tabular}{@{}c@{}}indian-liver\\patient\end{tabular} & -  & $0.51$ & $0.33$ & $0.57$  \\ 
  % \hline
 \begin{tabular}{@{}c@{}}travel\\insurance\end{tabular} & $0.18$ & $0.17$ & $0.16$ & $0.18$ \\
 \midrule
 \textbf{Average} & $0.14$ & $0.22$  & $0.21$ & $0.24$ \\
\bottomrule
\end{tabular}
}
\end{center}
\caption{The average learned explanation weight for explanations w/wo quantifiers and explanations w/wo conjunctions of label aggregators for each task. Empty values in the table indicate that the linguistic element was absent in the explanations for the corresponding dataset.
}
\vspace{-0.1in}
\label{tab:quantifier-conjunction_learned_accuracy_factor weight}
\end{table}

\begin{figure*}[!h]
    \centering
         \begin{subfigure}[b]{0.45\textwidth}
             \centering
             \includegraphics[width=\textwidth]{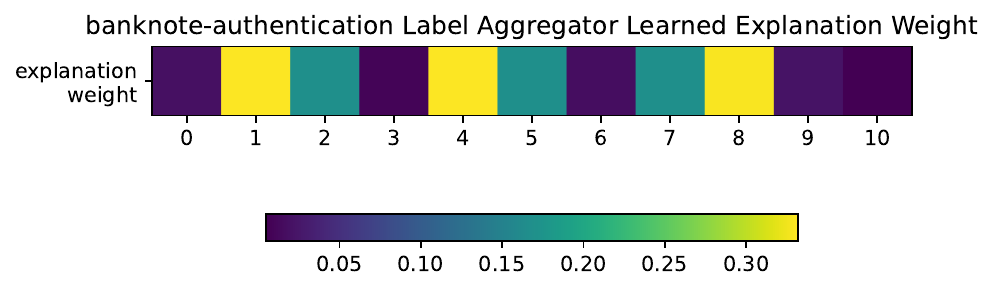}
             \caption{banknote-authentication}
             \label{fig:acc_ba}
         \end{subfigure}
         \hfill
         \begin{subfigure}[b]{0.45\textwidth}
             \centering
             \includegraphics[width=\textwidth]{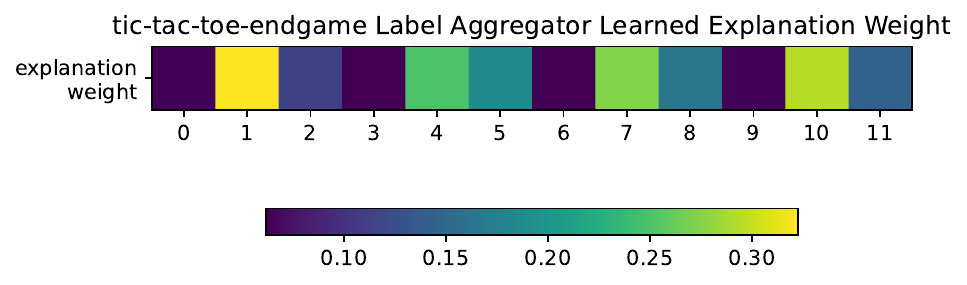}
             \caption{tic-tac-toe-endgame}
             \label{fig:acc_ttte}
         \end{subfigure}
         \hfill
         \begin{subfigure}[b]{0.45\textwidth}
             \centering
             \includegraphics[width=\textwidth]{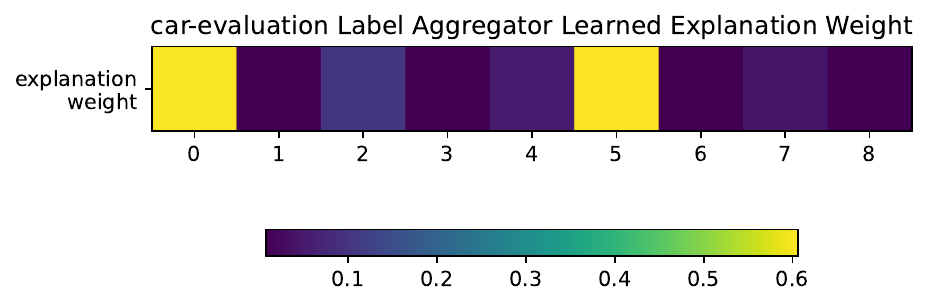}
             \caption{car-evaluation}
             \label{fig:acc_ce}
         \end{subfigure}
         \hfill
         \begin{subfigure}[b]{0.45\textwidth}
             \centering
             \includegraphics[width=\textwidth]{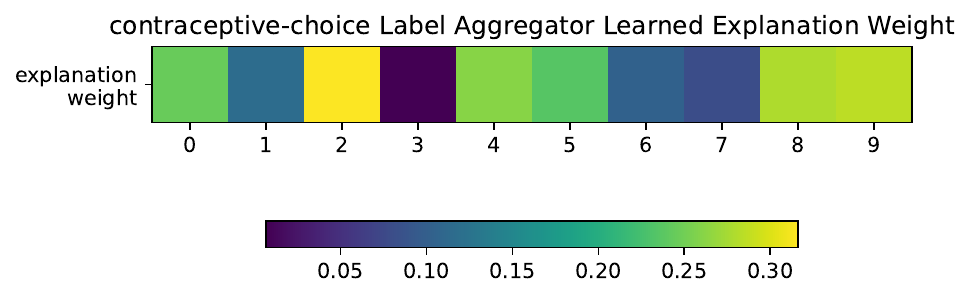}
             \caption{contraceptive-method-choice}
             \label{fig:acc_cmc}
         \end{subfigure}
         \hfill
         \begin{subfigure}[b]{0.45\textwidth}
             \centering
             \includegraphics[width=\textwidth]{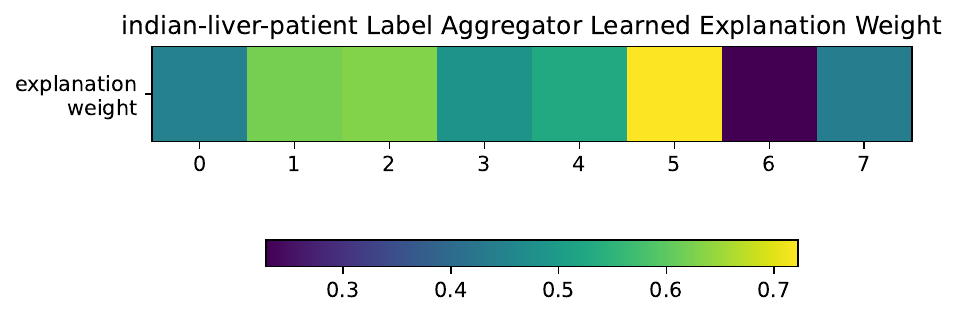}
             \caption{indian-liver-patient}
             \label{fig:acc_ilp}
         \end{subfigure}
         \hfill
         \begin{subfigure}[b]{0.45\textwidth}
             \centering
             \includegraphics[width=\textwidth]{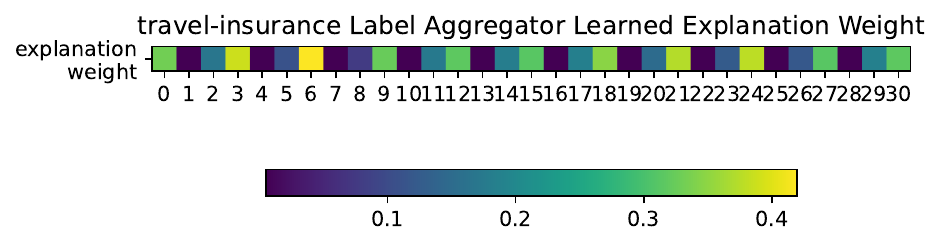}
             \caption{travel-insurance}
             \label{fig:acc_ilp}
         \end{subfigure}
    \caption{Learned label aggregator accuracy factor for the 6 \cluesreal evaluation datasets used in our work.}
    \label{fig:other learned weight}
\end{figure*}

\subsection{Individual Explanation for Each Task} \label{sec:app_nlexp}
\begin{figure*}[!h]
    \centering
    \begin{subfigure}[b]{0.45\textwidth}
    \centering
    \includegraphics[scale=0.45]{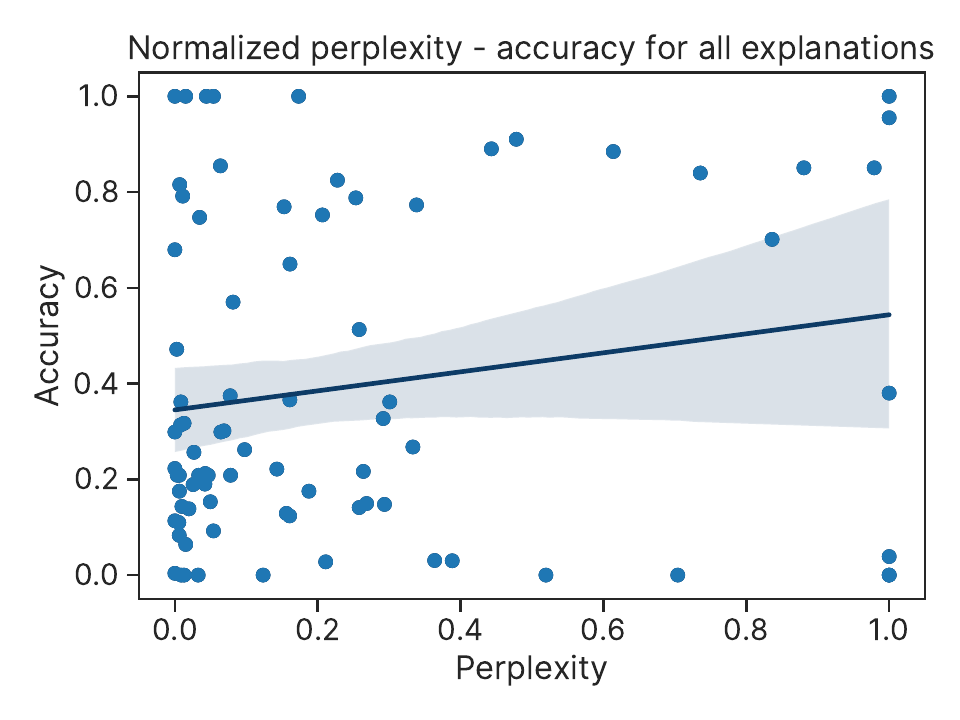}
   \vspace{-0.3cm}
    \caption{Pearson correlation coefficient = 0.18.}
    \end{subfigure}
    \hfill
    \begin{subfigure}[b]{0.45\textwidth}
        \centering
        \includegraphics[scale=0.45]{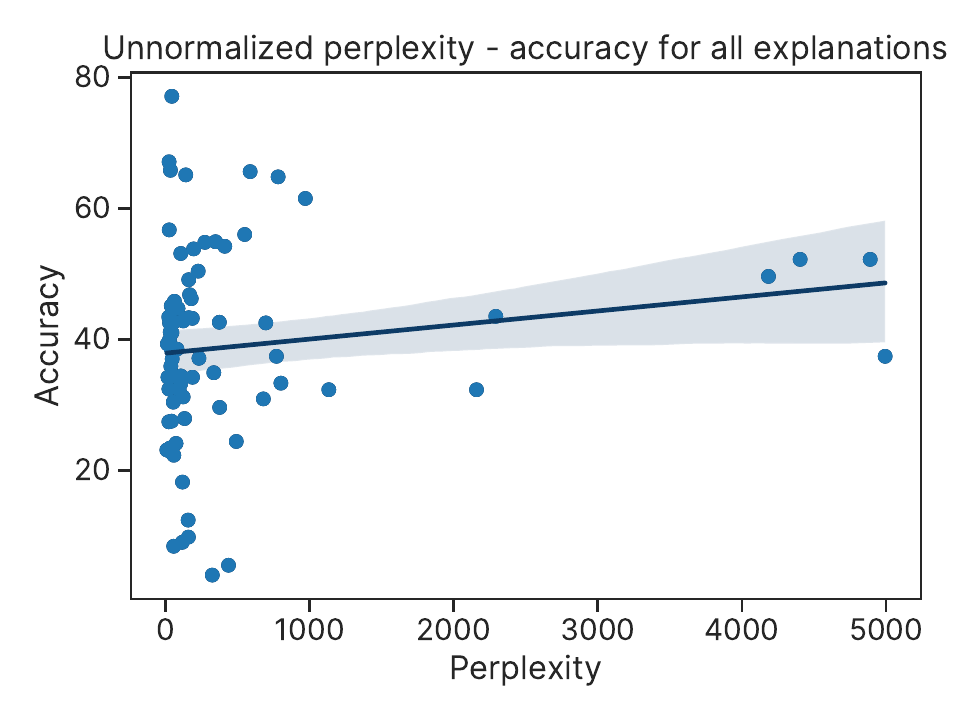}
       \vspace{-0.3cm}
        \caption{Pearson correlation coefficient = 0.15.}
    \end{subfigure}
    \caption{Relationship between perplexity and accuracy of explanations: (a) Normalized, (b) Un-normalized}
    \label{fig:accuracy-perplexity}
\end{figure*}

We show all the available natural language explanations for the six \cluesreal dataset we use in this paper in Table~\ref{table: banknote-authentication all explanations} to~\ref{table: travel-insurance all explanations}. In Table~\ref{table: banknote-authentication all explanations} to~\ref{table: travel-insurance all explanations}, We also report the accuracy when using only one explanation at a time with \exent and the perplexity of each explanations. In Figure~\ref{fig:accuracy-perplexity}, we analyze the correlation between accuracy and perplexity of all explanations. There is a positive correlation between accuracy and perplexity of all the explanations.

\subsection{Results for Models Without Abstention} \label{sec:app_noabst_results}
\begin{figure*}[!t]
    \begin{center}
        \includegraphics[scale=0.41]{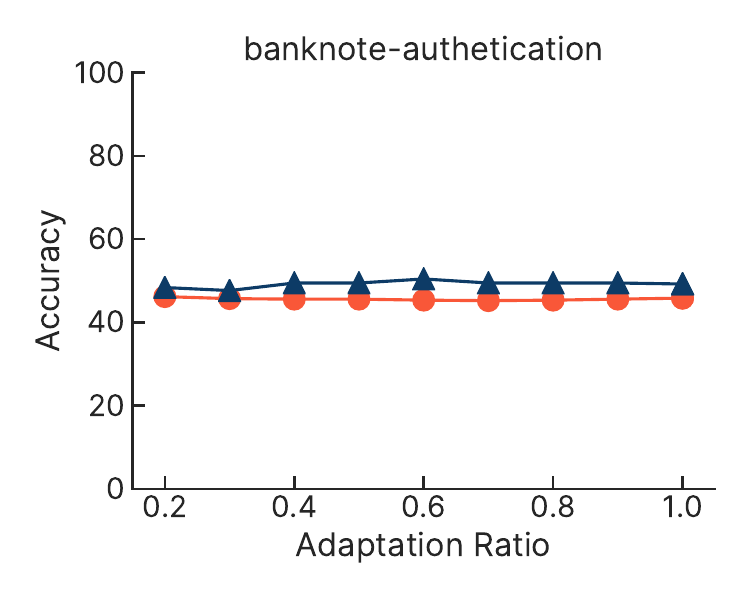}
        \includegraphics[scale=0.41]{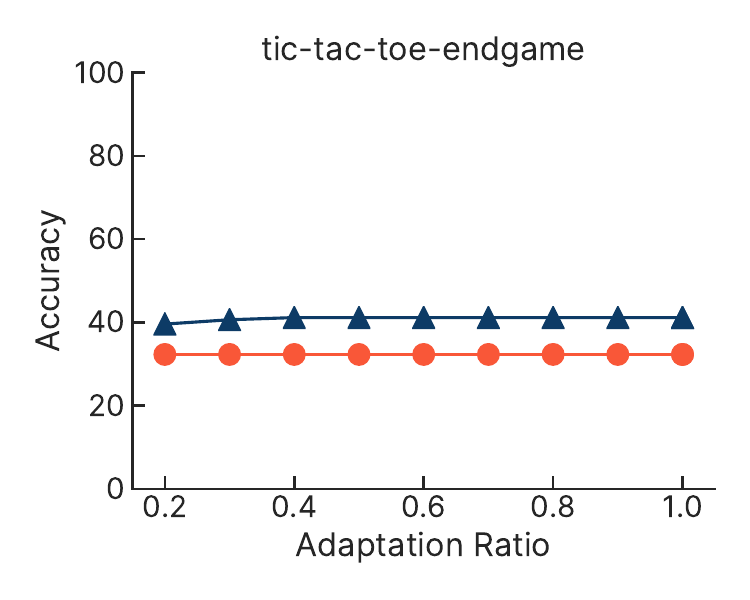}
        \includegraphics[scale=0.41]{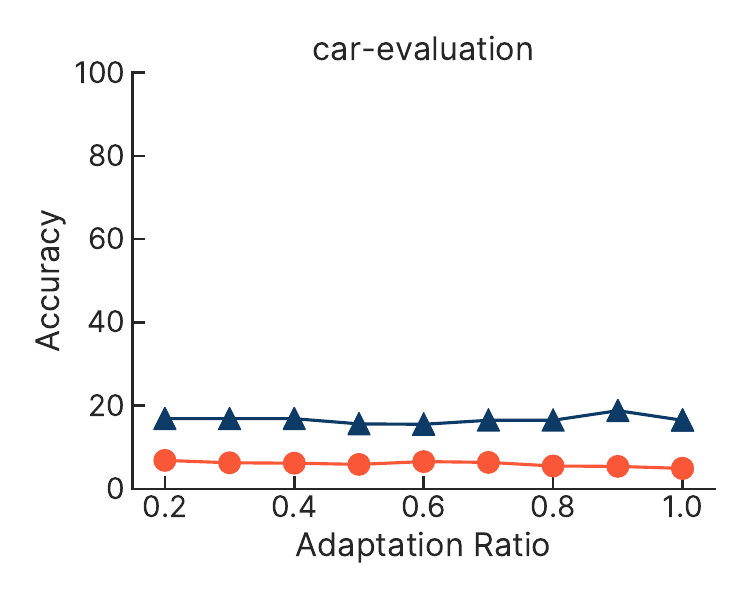}
        \includegraphics[scale=0.41]{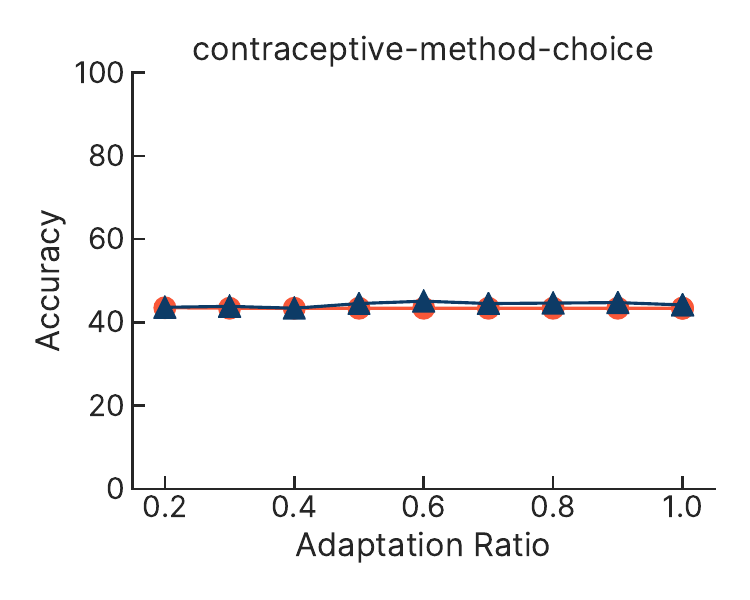}
        \includegraphics[scale=0.41]{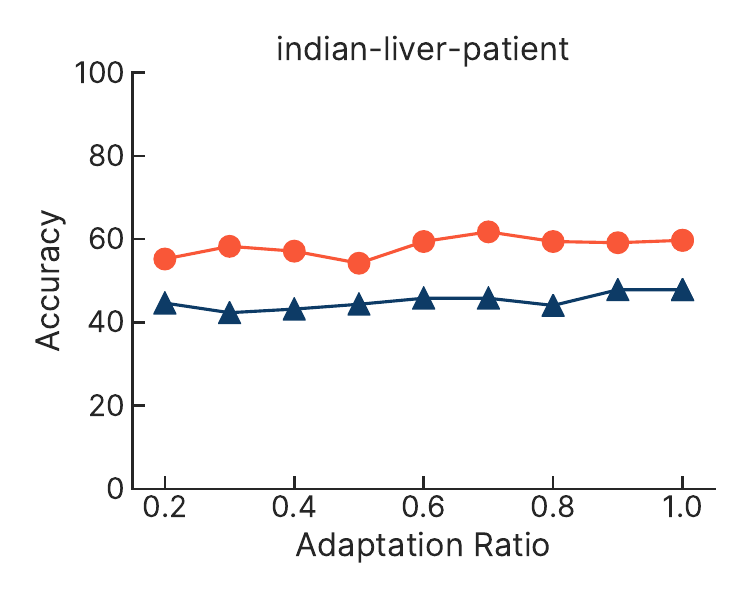}
        \includegraphics[scale=0.41]{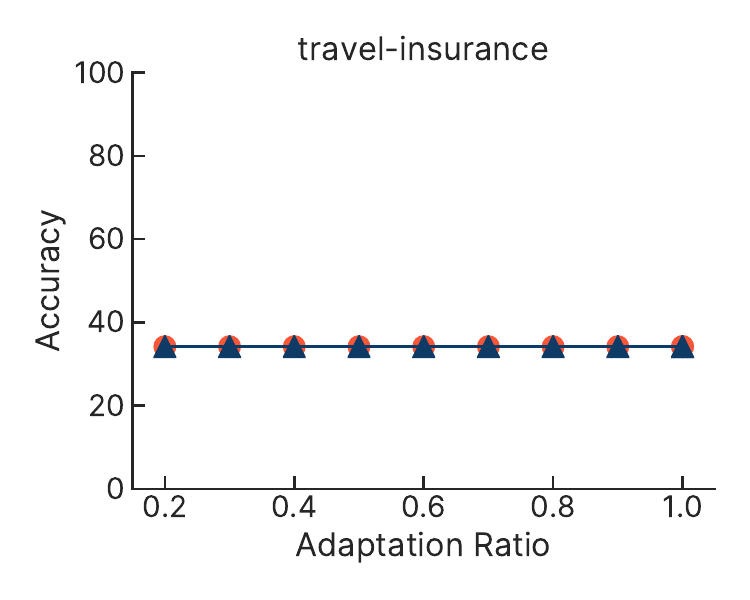}
        \includegraphics[scale=0.41]{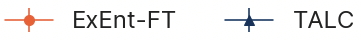}
    \end{center}
    \caption{Performance of \exent-FT and \nalla with different adaptation ratios for each of the 6 evaluation tasks. }
    \label{fig:different data size each task}
\end{figure*}

Here, we show \exent-FT and \nalla experiment results without abstention on different adaptation size for each 6 tasks from \cluesreal in Figure~\ref{fig:different data size each task}.

\subsection{Few Shot Learning}
\label{sec:few shot}
Here, we run a few-shot supervised version of \exent. We fine-tune the \exent model using $k$ samples with gold labels from the evaluation tasks, where $k=4, 8, 16, 32$. We report the results in Table~\ref{tab:few shot}. We observed that the test accuracy of this few-shot trained \exent is better than \nalla. The performance with a few-shot model is better as the gold labels are quite different from noisy aggregated labels used by \nalla for adaptation. We observed a huge label imbalance in the intermediate \exent model that results in lower accuracy for both \nalla and \exent-FT, both of which leverage \exent’s predictions as noisy pseudo labels.

\begin{table}[h!]
\begin{center}
\scalebox{0.75}{
\begin{tabular}{crrrr} 
 \toprule
\textbf{\textbf{Dataset}} & \textbf{$k=4$} & \begin{tabular}{@{}c@{}}\textbf{$k=8$}\end{tabular} 
& \begin{tabular}{@{}c@{}}\textbf{$k=16$}\end{tabular} & \textbf{$k=32$} \\
 \midrule
 \begin{tabular}{@{}c@{}}banknote\\authentication\end{tabular}  & $47.5_{(2.7)}$ & $51.2_{(1.5)}$  & $54.9_{(5.2)}$ & $53.6_{(4.1)}$ \\
 \begin{tabular}{@{}c@{}}tic-tac-toe\\endgame\end{tabular} & $55.6_{(8.1)}$ & $59.1_{(6.5)}$ & $63.7_{(2.8)}$ & $64.2_{(1.8)}$  \\ 
 \begin{tabular}{@{}c@{}}car\\evaluation\end{tabular} & $61.7_{(6.2)}$  & $52.6_{(1.6)}$  & $68.1_{(1.4)}$ & $69.4_{(2.3)}$ \\ 
\begin{tabular}{@{}c@{}}contraceptive\\choice\end{tabular} & $34.4_{(4.3)}$ & $32.9_{(4.4)}$ & $36.8_{(0.2)}$ & $40.0_{(2.3)}$ \\ 
 \begin{tabular}{@{}c@{}}indian-liver\\patient\end{tabular} & $66.9_{(2.6)}$ & $67.4_{(1.3)}$ & $63.8_{(3.4)}$ & $67.5_{(0.8)}$\\ 
 \begin{tabular}{@{}c@{}}travel\\insurance\end{tabular} & $53.9_{(10.3)}$ & $57.0_{(7.1)}$ & $53.9_{(1.7)}$ & $62.5_{(3.7)}$ \\
 \midrule
 \textbf{Average} & $53.3$ & $53.4$  & $56.9$ & $59.5$ \\
\bottomrule
\end{tabular}
}
\end{center}
\caption{Few-shot fine-tuning with \exent. We report the mean and standard deviation for the accuracy across three runs using different seeds.
}
\vspace{-0.1in}
\label{tab:few shot}
\end{table}

We would like to emphasize that a few-shot supervised model is \textbf{not} an ideal baseline for our framework. This is because \nalla is designed to work as an unsupervised approach for test-time adaptation. Hence, few-shot fine-tuning would represent an upper bound for \nalla. Alternatively, few-shot finetuning can be treated as a complementary approach to \nalla and can be paired with \nalla for real-world test time adaptation scenarios. Combining these two paradigms for improved test time adaptation is an interesting direction for future work and is beyond the scope of this paper.

\subsection{Ranking by Perplexity}
\label{sec:perplexity}
\begin{figure}[!h]
    \centering
    \includegraphics[scale=0.49]{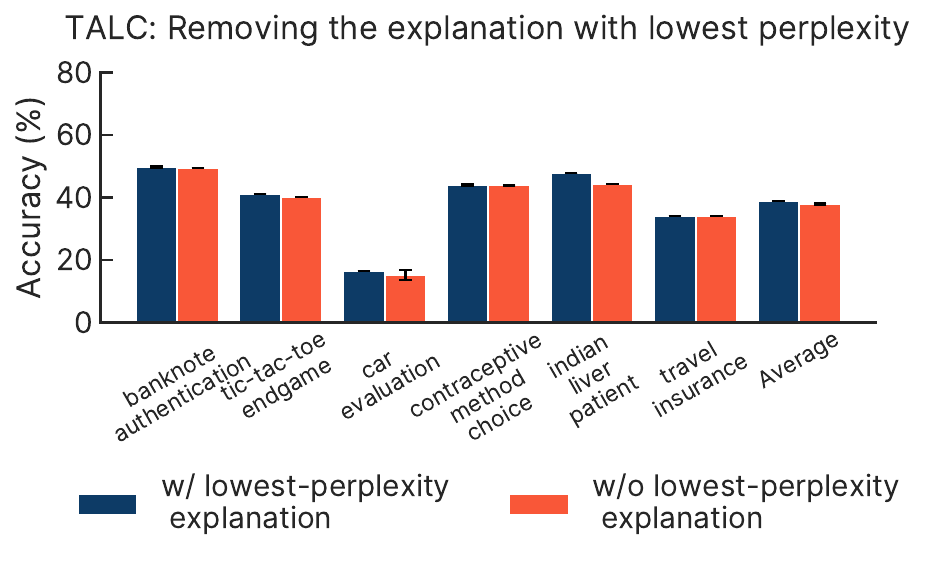}
    \caption{When ranking the explanations by their individual perplexity, removing the best explanation leads to a $1.0\%$ drop in performance on average.
    }
    \label{fig:gac no best exp p}
\end{figure}

\begin{figure}[!h]
    \centering
    \includegraphics[scale=0.5]{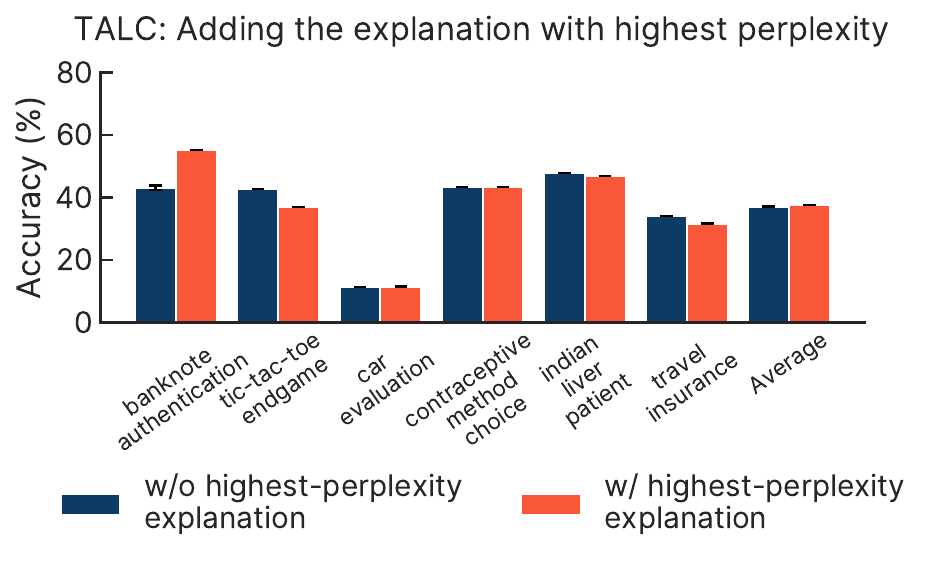}
    \caption{
    Comparison of \nalla's performance before and after adding a low-quality explanation to a set of high-quality explanations. On average, the performance increases by $0.6\%$ when ranking by perplexity.
    }
    \label{fig:gac one bad exp p}
\end{figure}

\begin{figure}[!h]%
    \centering
    \includegraphics[scale=0.5]{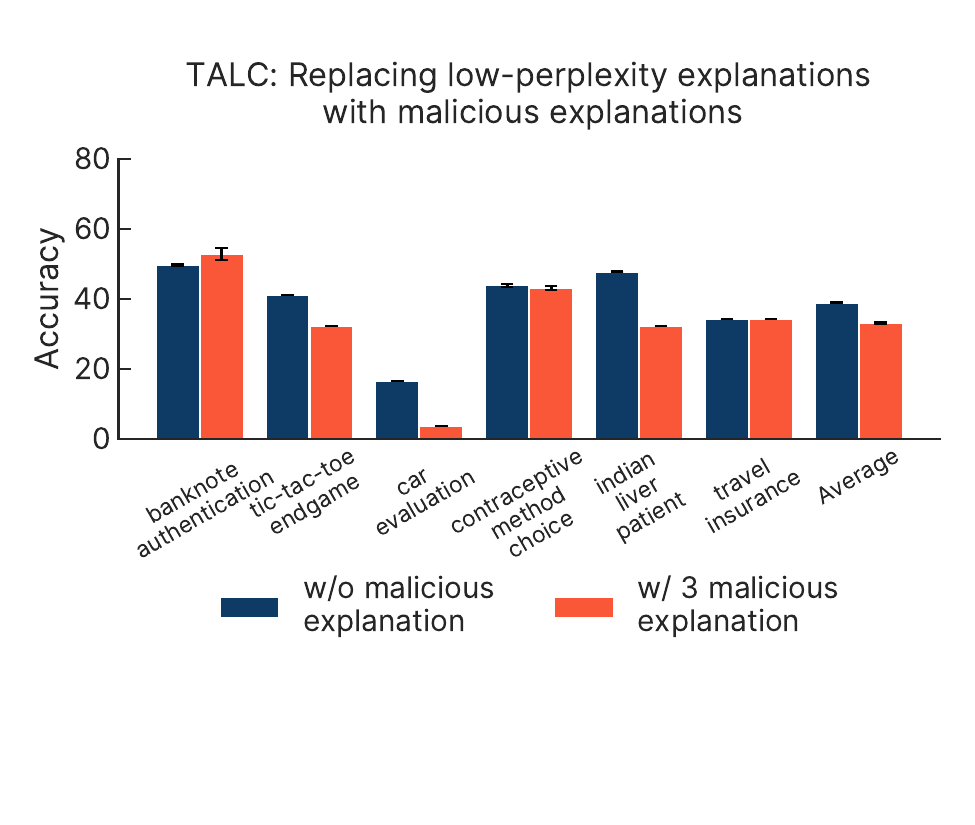}
    \caption{
    Comparison of \nalla's performance before and after replacing good-quality explanations with malicious explanations. On average, \nalla's performance drops by $5.6\%$ as good explanations are replaced by malicious explanations.
    }
    \label{fig:malicious exp p}
\end{figure}

We show the results of the ablations studies described in Section~\ref{sec:analysis} ranking by perplexity here in Figure~\ref{fig:gac no best exp p} to \ref{fig:malicious exp p}.

For the setting where we add one low-quality explanation to a set of high-quality explanation, when ranking by perplexity, the average performance increases by $0.6\%$, caused by the performance increase of the banknote-authentication task while the other tasks' performance either decreases or stays the same. According to Table~\ref{table: banknote-authentication all explanations}, the added explanation for banknote-authentication has the highest perplexity and the second highest accuracy, suggesting the accuracy metric may have stronger impact to the performance than perplexity does. 

\begin{table*}[ht]
\centering
\scalebox{1.0}{
 \begin{tabular}{c|p{9.5cm}|c|c} 
 \toprule
 \Thead{No.} & \Thead{Explanation} & \Thead{Accuracy}& \Thead{Perplexity}\\
 \midrule
 0 & If the variance of the note is a negative number, it's more likely to be an original note. & 56.7\% & 25.05\\
 \midrule
 1 & If the kurtosis of the bill is a negative number, it's more likely to be a fake note. & 41.1\% & 31.82\\
 \midrule
 2 & A fake banknote has a variance level over 4.0. & 46.2\% & 177.85\\
 \midrule
 3 & Notes that have a negative Variance level will also have a negative Entropy level. & 45.1\% & 39.02\\
 \midrule
 4 & Notes that have a positive Skewness level will have a negative Kurtosis level. & 45.8\% & 61.06\\
 \midrule
 5 & Variance above 4 leads to the results of FAKE. & 43.3\% & 160.22\\
 \midrule
 6 & Below 3.80 skewness leads to the original. & 56.0\% & 548.93\\
 \midrule
 7 & Variance above 1.00 leads to the FAKE. & 53.1\% & 105.15\\
 \midrule
 8 & Below the 3.75 Sketwness leads to the ORIGINAL. & 54.2\% & 410.40\\
 \midrule
 9 & Entropy is low value so it is fake. & 49.1\% & 160.25\\
 \midrule
 10 & Kurtosis is high value so it is original. & 54.9\% & 346.58\\
 \midrule
 \end{tabular}}
 \caption{All explanations for banknote-authentication task used in this paper.}
 \label{table: banknote-authentication all explanations}
\end{table*}

\begin{table*}[ht]
\centering
\scalebox{1.}{
 \begin{tabular}{c|p{9.5cm}|c|c} 
 \toprule
 \Thead{No.} & \Thead{Explanation} & \Thead{Accuracy} & \Thead{Perplexity}\\
 \midrule
 0 & The game is usually not over yet when there are at least two blank squares. & 34.4\% & 56.82\\
 \midrule
 1 & The game is usually over when the players have taken all the middle squares. & 42.7\% & 51.84\\
 \midrule
 2 & The top middle square results in x winning. & 33.3\% & 800.38\\
 \midrule
 3 & The bottom middle square results in x winning. & 32.3\% & 1132.84\\
 \midrule
 4 & When the middle-right-square is left blank, x is less likely to win. & 32.3\% & 94.01\\
 \midrule
 5 & Whoever marks the top-left-square is unlikely to win. & 38.5\% & 78.83\\
 \midrule
 6 & The player who moves second is much more likely to lose if they place in the middle-left-square. & 37.0\% & 45.58\\
 \midrule
 7 & The middle-right-square is the square that is most likely to go unused during a game. & 35.9\% & 36.20\\
 \midrule
 8 & Top left square X indicates the Negative group. & 61.5\% & 969.94\\
 \midrule
 9 & Without b categories in middle middle square comes under the Positive group. & 32.3\% & 2158.20\\
 \midrule
 10 & An O in both the top-left and bottom-right is likely to be positive. & 39.6\% & 24.43\\
 \midrule
 11 & A blank in the middle-right will likely lead to negative. & 65.1\% & 139.76\\
 \midrule
 \end{tabular}}
 \caption{All explanations for tic-tac-toe-endgame task used in this paper.}
 \label{table: tic-tac-toe-endgame all explanations}
\end{table*}

\begin{table*}[ht]
\centering
\scalebox{1.}{
 \begin{tabular}{c|p{9.5cm}|c|c} 
 \toprule
 \Thead{No.} & \Thead{Explanation} & \Thead{Accuracy} & \Thead{Perplexity}\\
 \midrule
 0 & If safety is high, then the car will not be unacceptable. & 22.3\% & 56.95\\
 \midrule
 1 & If maintenance cost is medium, then the car will not be unacceptable. & 42.8\% & 121.90\\
 \midrule
 2 & A capicity for 4 or more persons makes the vehicle acceptable for resale. & 9.0\% & 115.36\\
 \midrule
 3 & High safety ratings generally make vehicles acceptable for resale. & 9.8\% & 158.10\\
 \midrule
 4 & A low buying cost is a good indicator of a vehicle being acceptable for resale. & 8.4\% & 55.96\\
 \midrule
 5 & Most people having the passenger capacity of 4 or more have good acceptability for car resale. & 18.2\% & 117.22\\
 \midrule
 6 & Cars with low buying and maintenance cost  are highly acceptable for resale & 5.5\% & 436.58\\
 \midrule
 7 & Cars with higher safety and capacity are highly acceptable for resale. & 12.4\% & 156.38\\
 \midrule
 8 & Cars with higher safety and medium luggage boot size are highly acceptable for resale. & 4.0\% & 323.92\\
 \midrule
 \end{tabular}}
 \caption{All explanations for car-evaluation task used in this paper.}
 \label{table: car-evaluation all explanations}
\end{table*}

\begin{table*}[ht]
\centering
\scalebox{1.}{
 \begin{tabular}{c|p{9.5cm}|c|c} 
 \toprule
 \Thead{No.} & \Thead{Explanation} & \Thead{Accuracy} & \Thead{Perplexity}\\
 \midrule
 0 & A husband's education has a high chance of a long-term contraceptive being used. & 44.7\% & 85.88\\
 \midrule
 1 & A wife's education level usually determines if a long term contraceptive was used. & 46.8\% & 166.37\\
 \midrule
 2 & Women with low education are more likely to use short-term contraception. & 39.3\% & 12.18\\
 \midrule
 3 & Women with high education age 40 and under are more likely to use long-term contraception. & 43.4\% & 22.01\\
 \midrule
 4 & Women with two or more children used short and long term methods. & 42.7\% & 47.27\\
 \midrule
 5 & The least educated women either used short-term method or didn't use any contraceptive method. & 24.1\% & 72.05\\
 \midrule
 6 & If the wife's education is high, then the contraceptive method used is long-term. & 41.0\% & 44.04\\
 \midrule
 7 & If the wife's education is not high, then the contraceptive method used is no-use or short-term. & 27.5\% & 41.12\\
 \midrule
 8 & If the wife's education is not high, then the contraceptive method used is not long-term. & 23.4\% & 31.23\\
 \midrule
 9 & If the wife's education is not high, then the contraceptive method is short-term. & 38.6\% & 37.03\\
 \midrule
 \end{tabular}}
 \caption{All explanations for contraceptive-method-choice task used in this paper.}
 \label{table: contraceptive-method-choice  all explanations}
\end{table*}

\begin{table*}[ht]
\centering
\scalebox{1.}{
 \begin{tabular}{c|p{9.5cm}|c|c} 
 \toprule
 \Thead{No.} & \Thead{Explanation} &\Thead{Accuracy} & \Thead{Perplexity}\\
 \midrule
 0 & The SGPT High percentage so the liver patient was yes. & 49.6\% & 4184.78\\
 \midrule
 1 & The SGPT Low percentage so the liver patient was no. & 52.2\% & 4891.02\\
 \midrule
 2 & Patients over the age of forty are liver patients. & 42.6\% & 54.92\\
 \midrule
 3 & Patients who has SGOT range greater than forty are liver patients. & 42.6\% & 373.03\\
 \midrule
 4 & Decreased SGPT Value ensures no liver patient. & 52.2\% & 4404.52\\
 \midrule
 5 & Age group above 40 ensures liver patient. & 37.4\% & 4994.60\\
 \midrule
 6 & Some people have more age and have the SGOT and they are liver patient. & 54.8\% & 272.55\\
 \midrule
 7 & Some people have minimum age and they are liver patient. They are somewhat accurate. & 50.4\% & 226.11\\
 \midrule
 \end{tabular}}
 \caption{All explanations for indian-liver-patient task used in this paper.}
 \label{table: indian-liver-patient all explanations}
\end{table*}

\begin{table*}[ht]
\centering
\scalebox{0.8}{
 \begin{tabular}{c|p{12cm}|c|c} 
 \toprule
 \Thead{No.} & \Thead{Explanation} & \Thead{Accuracy} & \Thead{Perplexity}\\
 \midrule
 0 & Frequent flyers with an annual income over 1 million usually take travel insurance. & 33.2\% & 104.36\\
 \midrule
 1 & People with an annual income over 1 million and under 30 years old usually take travel insurance. & 30.4\% & 53.40\\
 \midrule
 2 & Travelers who are 29 and older take traveler insurance. & 34.4\% & 105.36\\
 \midrule
 3 & Travelers who have not traveled abroad before are more likely to take traveler insurance. & 34.2\% & 22.74\\
 \midrule
 4 & Frequent flyer travelers with an annual income above 1 million frequently take travel insurance. & 31.2\% & 121.70\\
 \midrule
 5 & Travelers older than 25 years old and with an income below 1 million do not usually take travel insurance. & 42.5\% & 27.45\\
 \midrule
 6 & Most passengers who are not frequent fliers do not use travel insurance. & 65.8\% 33.28\\
 \midrule
 7 & Most passengers who have not traveled abroad do not use travel insurance. & 77.1\% & 42.87\\
 \midrule
 8 & Most passenger with an income higher than 100,000 use travel insurance. & 27.9\% & 131.69\\
 \midrule
 9 & People who have never traveled abroad before are more likely to have taken travel insurance. & 34.2\% & 15.95\\
 \midrule
 10 & People with an annual income below 1,000,000 are less likely to have traveled abroad than those with annual incomes above 1,000,000. & 23.1\% & 8.36\\
 \midrule
 11 & People with an annual income above 1,000,000 are more likely to have taken travel insurance. & 39.9\% & 26.63\\
 \midrule
 12 & More frequent flyers have taken travel insurance. & 34.9\% & 334.32\\
 \midrule
 13 & More people that travel abroad have taken travel insurance. & 34.2\% & 114.91\\
 \midrule
 14 & People who are non-frequent flyers and are college graduates are less likely to get travel insurance. & 27.4\% & 22.28\\
 \midrule
 15 & People who make a million or more and are frequent fliers are more likely to get travel insurance. & 32.4\% & 22.73\\
 \midrule
 16 & Most people who didn't travel abroad before had taken Travel Insurance. & 34.2\% & 84.50\\
 \midrule
 17 & Most college graduate have taken Travel Insurance. & 64.8\% & 780.98\\
 \midrule
 18 & Those who are college graduates and in their 20s are somewhat likely to purchase travel insurance. & 22.9\% & 28.62\\
 \midrule
 19 & Those who have never travelled abroad and are not frequent flyers often do not purchase travel insurance. & 67.1\% & 23.85\\
 \midrule
 20 & Travelled Abroad Before "No" categories indicates the "No" Travel Insurance Taken. & 24.4\% & 490.45\\
 \midrule
 21 & College Graduate "Yes" categories leads to the "Yes"  Travel Insurance Taken. & 30.9\% & 678.44\\
 \midrule
 22 & Frequent Flyer "No" indicates the "No" Travel Insurance Taken. & 53.8\% & 194.31\\
 \midrule
 23 & Annual income categories above  1049999 indicates the "Yes" Travel Insurance Taken. & 43.5\% & 2291.89\\
 \midrule
 24 & Most college graduates that make more than 1000000 annually have taken travel insurance. & 37.2\% & 231.38\\
 \midrule
 25 & Most college graduates from 26 to 34 have taken travel insurance. & 43.2\% & 185.20\\
 \midrule
 26 & About have of college graduates that have not travelled abroad before have taken travel insurance. & 34.2\% & 186.52\\
 \midrule
 27 & Frequently flyer "No" categories indicates the "No" travel insurance taken group. & 65.6\% & 586.77\\
 \midrule
 28 & Annual income above 1049999 indicate the "Yes" travel insurance taken group. & 37.4\% & 769.28\\
 \midrule
 29 & Frequent flyer "No" categories indicates the "No" travel insurance taken group. & 29.6\% & 375.25\\
 \midrule
 30 & Annual income above 1049999 indicates the "Yes" travel insurance taken group. & 42.5\% & 695.34\\
 \midrule
 \end{tabular}}
 \caption{All explanations for travel-insurance task used in this paper.}
 \label{table: travel-insurance all explanations}
\end{table*}

\end{document}